\def\year{2022}\relax
\documentclass[letterpaper]{article} 
\usepackage{aaai22}  
\usepackage{times}  
\usepackage{helvet}  
\usepackage{courier}  
\usepackage[hyphens]{url}  
\usepackage{graphicx} 
\usepackage{natbib}  
\usepackage{caption} 
\DeclareCaptionStyle{ruled}{labelfont=normalfont,labelsep=colon,strut=off} 
\usepackage{algorithm}
\usepackage[noend]{algpseudocode}
\usepackage{subcaption}
\usepackage{caption}
\usepackage{booktabs} 
\usepackage{stmaryrd} 
\usepackage{xcolor}
\usepackage{mathtools}
\usepackage{amsfonts}
\usepackage{amsthm}
\usepackage[hidelinks]{hyperref}

\theoremstyle{definition}
\newtheorem{definition}{Definition}
\newtheorem{subdefinition}{Definition}[definition]

\usepackage{color} 

\newtheorem{theorem}{Theorem}[section]
\newtheorem{lemma}[theorem]{Lemma}

\title{Solving Disjunctive Temporal Networks with Uncertainty under Restricted Time-Based Controllability using Tree Search and Graph Neural Networks}

\author{
    \parbox{\linewidth}{\centering
        Kevin Osanlou\textsuperscript{1}, 
        Jeremy Frank\textsuperscript{1}, 
        Andrei Bursuc\textsuperscript{2}, 
        Tristan~Cazenave\textsuperscript{3}, 
        Eric Jacopin\textsuperscript{4}, 
        Christophe Guettier\textsuperscript{5} and 
        J. Benton\textsuperscript{1}
    } \\
    \vspace{0.3cm}
    {\fontsize{10}{10} \normalfont \selectfont$^1$ NASA Ames Research Center, $^2$ Valeo.ai, $^3$LAMSADE, Paris-Dauphine University,
    $^4$CREC~Saint-Cyr~Coetquidan, $^5$ Safran \\
    \vspace{0.1cm}
    {\{kevin.osanlou, jeremy.d.frank, j.benton\}@nasa.gov}, {andrei.bursuc@valeo.com},
    {tristan.cazenave@lamsade.dauphine.fr}, {christophe.guettier@safrangroup.com},
	 {eric.jacopin@st-cyr.terre-net.defense.gouv.fr}}
}

\begin{document}

\maketitle

\def\fromjeremy#1{{\color{blue}\small{\bf JF:} {\em #1}}}
\def\fromkevin#1{{\color{red}\small{\bf KO:} {\em #1}}}
\def\fromJ#1{{\color{yellow}\small{\bf JB:} {\em #1}}}
\def\fromEric#1{{\color{pink}\small{\bf EJ:} {\em #1}}}
\def\fromTristan#1{{\color{purple}\small{\bf TC:} {\em #1}}}
\def\fromAndrei#1{{\color{brown}\small{\bf AB:} {\em #1}}}
\def\fromChristophe#1{{\color{green}\small{\bf CG:} {\em #1}}}

\newcommand{\ab}[1]{\textcolor{orange}{#1}}
\newcommand{\abc}[1]{\textcolor{orange}{[AB: #1]}}

\def\authassign#1{{\color{orange}\small{\bf Contributor(s):} {\em #1}}}
\def\dor{\textit{d-OR }}
\def\wor{\textit{w-OR }}
\def\and{\textit{AND }}
\def\wait{\textit{WAIT }}
\def\etal{\emph{et al.}}
\def\ie{\emph{i.e. }}
\def\eg{\emph{e.g. }}
\def\True{\emph{True }}
\def\true{\emph{true }}
\def\False{\emph{False }}
\def\false{\emph{false }}

\begin{abstract}
Planning under uncertainty is an area of interest in artificial intelligence. We present a novel approach based on tree search and graph machine learning for the scheduling problem known as Disjunctive Temporal Networks with Uncertainty (DTNU). Dynamic Controllability (DC) of DTNUs seeks a reactive scheduling strategy to satisfy temporal constraints in response to uncontrollable action durations. We introduce new semantics for reactive scheduling: Time-based Dynamic Controllability (TDC) and a restricted subset of TDC, R-TDC. We design a tree search algorithm to determine whether or not a DTNU is R-TDC. Moreover, we leverage a graph neural network as a heuristic for tree search guidance. Finally, we conduct experiments on a known benchmark on which we show R-TDC to retain significant completeness with regard to DC, while being faster to prove. This results in the tree search processing fifty percent more DTNU problems in R-TDC than the state-of-the-art DC solver does in DC with the same time budget. We also observe that graph neural network search guidance leads to substantial performance gains on benchmarks of more complex DTNUs, with up to eleven times more problems solved than the baseline tree search.
\end{abstract}

\section{Introduction and Related Works}

\begin{figure}[t!]
\centering
\includegraphics[scale=0.64]{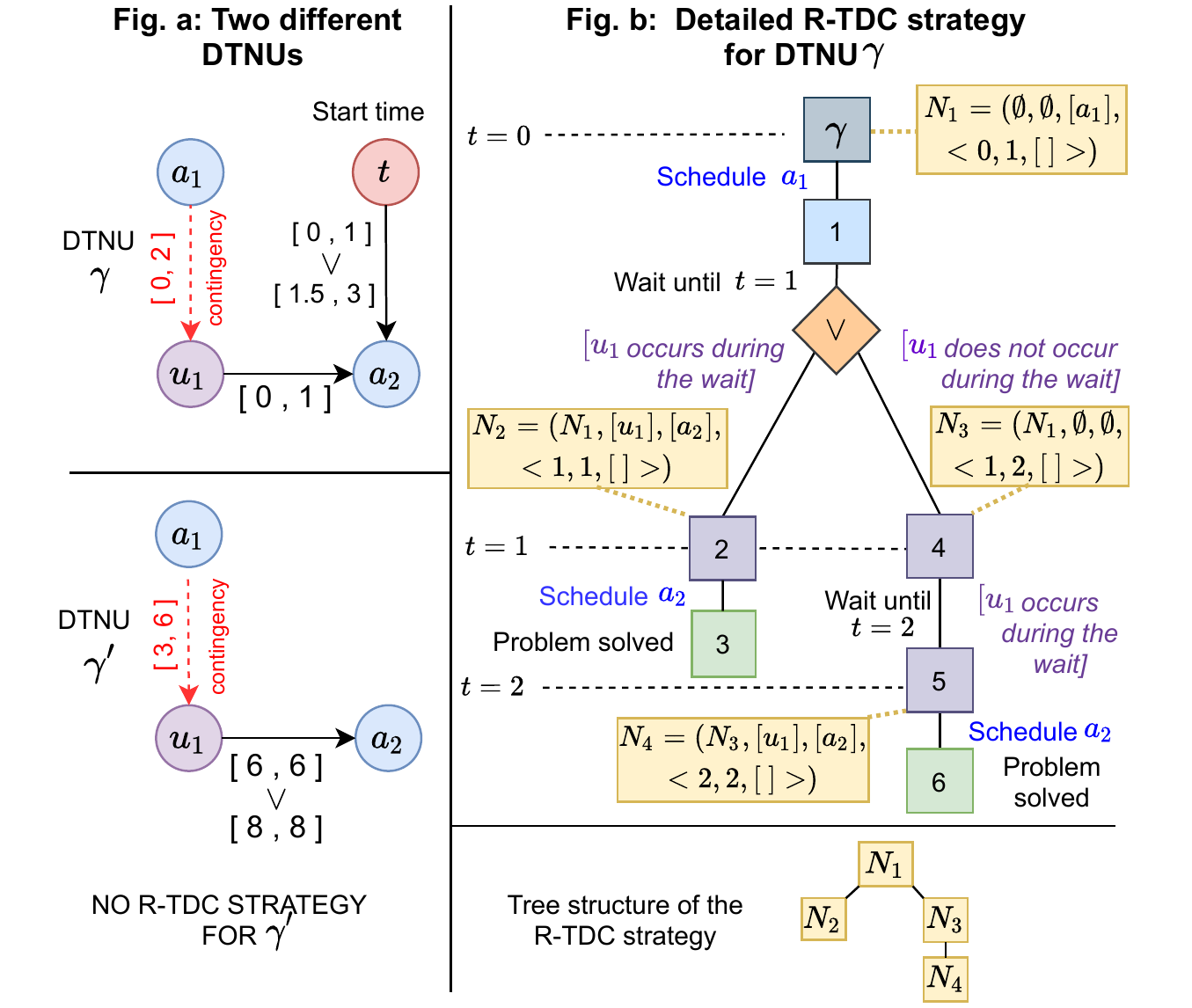}
\caption{\fontsize{10}{10} \normalfont \selectfont Two example DTNUs $\gamma$ and $\gamma'$. Timepoints $a_1$ and $a_2$ are controllable; $u_1$ uncontrollable. Black arrows represent time constraints between timepoints; red arrows contingency links. A detailed R-TDC strategy is displayed for $\gamma$. Squares below $\gamma$ are sub-DTNUs; the $\lor$ sign lists transitional possibilities. Nodes $N_i$ are R-TDC strategy nodes.
}
\label{fig:easy-example}
\vspace{-0.5cm}
\end{figure}

Temporal Networks (TN) are a common formalism to represent temporal constraints over a set of time points (\emph{e.g.} start/end of activities in a scheduling problem). The Simple Temporal Network with Uncertainty (STNUs) introduced by \citet{kn:ViFa}
explicitly incorporates qualitative uncertainty into temporal networks.  Applications include control of robotic systems such as in \citet{kn:BhMuVaWi} and \citet{kn:VaChAg}, with limited ethical concerns thus far. Considerable work has resulted in algorithms to determine whether or not all timepoints can be scheduled, either up-front or reactively, in order to account for uncertainty (\emph{e.g.} \citet{kn:MoMu2}, \citet{kn:Mofast}). Dynamic Controllability (DC) is a form of scheduling in which a controller agent integrates observed events as they unfold to adapt the scheduling reactively. In particular, an STNU is said to be DC if there is a reactive scheduling strategy in which controllable timepoints can be executed either at a specific time, or after observing the occurrence of an uncontrollable timepoint. \citet{cimatti2016dynamic} investigate the problem of DC for 
Disjunctive Temporal Networks with Uncertainty (DTNUs), which generalize STNUs.  
Figure~\ref{fig:easy-example}a
shows 
two DTNUs $\gamma$ and $\gamma'$ on the left side; $a_i$ are controllable timepoints, $u_j$ are uncontrollable timepoints. Timepoints are variables which can take on any value in $\mathbb{R}$. Constraints between timepoints characterize a minimum and maximum time distance separating them, likewise valued in $\mathbb{R}$.  The key difference between STNUs and DTNUs
lies in the {\em disjunctions} that yield more choice points for consistent scheduling, especially reactively (\eg if an uncontrollable timepoint occurs early, a given constraint could be satisfied, if it occurs late, another one, linked by a disjunction, could be).

DC-checking for STNUs is $\mathcal{O}(N^3)$ (\citet{kn:Mofast}), however, it is $PSPACE$-complete for DTNUs (\citet{kn:BhWi}), making this a highly challenging problem. The difficulty in proving or disproving DC arises from the need to check all possible combinations of disjuncts to handle all possible outcomes of uncontrollable timepoints. DTNUs are nonetheless more expressive than STNUs, and many real world applications require time-windows in which certain tasks can be scheduled either in a given interval or another, making it a problem worth studying. The best previously published approaches for this problem come from \citet{cimatti2016dynamic} and use timed-game automata and satisfiability modulo theories.

An emerging trend of neural networks, Graph Neural Networks (GNNs), has been proposed to extend convolutional neural networks (\citet{krizhevsky_2012}) to graph inputs for machine learning tasks. Recent variants based on spectral graph theory include works from \citet{defferrard_2016} and \citet{kipf_2017}. They leverage relational properties between nodes, but do not take into account potential edge weights. In newer spatial-based approaches, Message Passing Neural Networks (MPNNs) from \citet{battaglia2016interaction}, \citet{gilmer2017neural} and \citet{kipf2018neural} use embeddings comprising edge weights within each computational layer. We focus on these architectures as DTNUs can be formalized as graphs with edge distances representing time constraints.

In this work, we pose DC-checking of DTNUs as a search problem, express states as graphs, and use MPNNs to learn heuristics based on previously solved DTNUs to guide search.
The key contributions of our approach are the following. \textbf{(1)} We introduce new semantics for reactive scheduling: Time-based Dynamic Controllability (TDC), and a restricted subset of TDC, R-TDC. We present a tree search approach to identify R-TDC strategies. 
\textbf{(2)} We describe an MPNN architecture trained with self-supervised learning for handling DTNU scheduling problems and use it as heuristic for guidance in the tree search.
\textbf{(3)} 
We carry out experiments on a known benchmark showing that R-TDC retains significant completeness compared to DC while being faster to prove. This leads to $50$\% more DTNU instances processed in R-TDC by the tree search than in DC with the state-of-the-art DC solver in the same time budget. Moreover, we show that the learned MPNN heuristic considerably improves the tree search on benchmarks of harder DTNUs: performance gains go up to $11$ times more instances solved than the baseline tree search within the same time frame. Our results also highlight that the MPNN, which is trained on a set of solved DTNUs, is able to generalize to larger DTNUs than those on which it was trained.

\section{Problem and Controllability Definitions}
\label{tdc}

We next provide definitions necessary in the context of this work: Dynamic Controllability (DC), Time-based Dynamic Controllability (TDC) and Restricted TDC (R-TDC).

\begin{definition}[DTNU and variants]
A DTNU $\Gamma$ is a tuple \{A,U,C,L\}, where: A is a set of controllable timepoints; U a set of uncontrollable timepoints; C a set of free constraints, each of the form $\lor_{k=1}^{q} v_{k,j} - v_{k,i} \in [x_{k},y_{k}]$, for some $v_{k,j}, v_{k,i} \in V = A \cup U$, $x_k, y_k \in \mathbb{R} \cup \{-\infty, +\infty \}$ and $q \in \mathbb{Z}^+$; L a set of contingency links, each of the form $\langle a_i, \lor_{k=1}^{q'} [x_{k}',y_{k}'] ,u_j \rangle$ where $a_i\in A$, $u_j\in U$, $x_k', y_k' \in \mathbb{R} \cup \{-\infty, +\infty \}$, $0 \leq x_{k}' \leq y_{k}' \leq x_{k+1}' \leq y_{k+1}' ~ \forall k = 1, 2, ..., q'-1$ and $q' \in \mathbb{Z}^+$, indicating possible occurrence time intervals of $u_j$ after $a_i$. A DTNU without uncontrollable timepoints is referred to as Disjunctive Temporal Network (DTN). STNUs follow the same definition as DTNUs but do not contain any disjunction inside constraints. Finally, an STNU without uncontrollable timepoints is a Simple Temporal Network (STN).
\end{definition}

\begin{definition}[DC \& TDC]
\textbf{DC} is a reactive form of scheduling which incorporates occurrences of uncontrollable events as they unfold and adapts to them. A problem is DC if and only if it admits a valid dynamic strategy expressed as a map from partial schedules to Real-Time Execution Decisions (RTEDs) (\citet{cimatti2016dynamic}). A partial schedule represents the current scheduling state, \ie the set of timepoints that have been scheduled or occurred so far and their timing. RTEDs allow for two possible actions: \textbf{(1)} The wait action, \ie  wait for an uncontrollable timepoint to occur. \textbf{(2)} The $(t, \mathcal{X})$ action, \ie if nothing happens before time $t$, schedule the controllable timepoints in $\mathcal{X}$ at $t$. A strategy is valid if, for every possible occurrence of the uncontrollable timepoints, controllable timepoints get scheduled in a way that all free constraints are satisfied. A~\textbf{TDC} strategy is a representation of a DC strategy as a timed tree, \ie a map from tree nodes to children nodes.
Tree nodes represent partial schedules, and their children lead to the execution of one of the following actions: \textbf{(1)} Schedule a set of controllable timepoints at current time; \textbf{(2)} Wait a period of time or until an uncontrollable timepoint occurs, whichever happens first.
\end{definition}


\begin{definition}[R-TDC]
 \textbf{R-TDC} is a finite subset of TDC. In particular, actions associated to a partial schedule in R-TDC are: \textbf{(1)} Schedule a set of controllable timepoints at current time; \textbf{(2)} Wait an \textbf{uninterruptible} period of time, the wait duration being defined by time discretization rules in \S~\ref{waitperiod}.

\end{definition}

\begin{figure}[tb]
\centering
\includegraphics[scale=0.9]{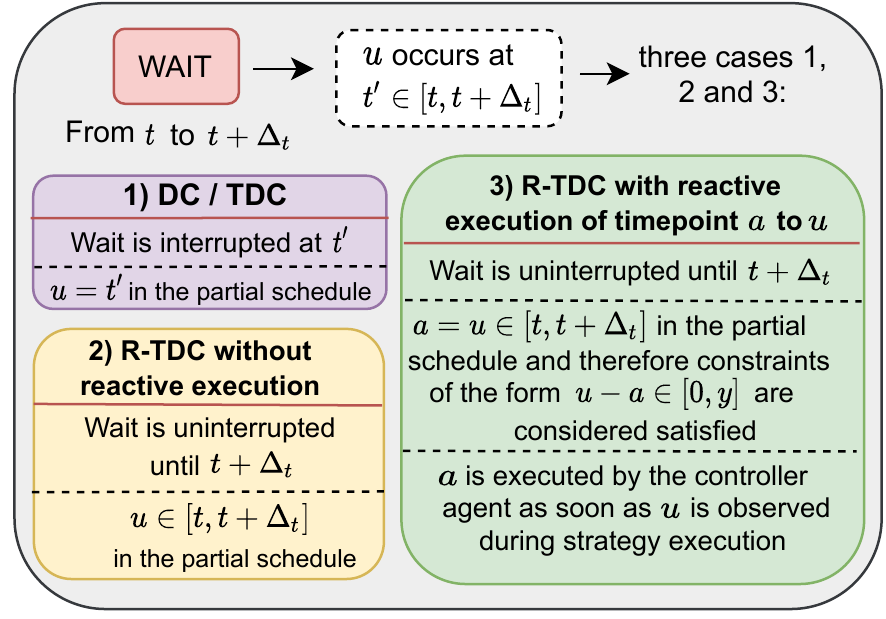}
\caption{\fontsize{10}{10} \normalfont \selectfont Execution of waits by controllability type. Timepoint $u$ is uncontrollable; $a$ is controllable. A wait from $t$ to $t+ \Delta_t$ is considered and corresponding behaviors are shown.
}
\label{fig:dctypes}
\vspace{-0.4cm}
\end{figure}

A TDC strategy can fully express a DC strategy which has an infinite number of mappings from partial schedules to RTEDs, given an infinite tree. In this work, we restrict TDC to a finite search space, R-TDC, and weigh how loss of search completeness results in increased efficiency. The restrictions in R-TDC stem from the uninterruptible waits: occurrence times of uncontrollable timepoints happening in waits are only bounded. Thus, partial schedules in R-TDC tree nodes do not carry exact occurrence times for uncontrollable timepoints which already occurred but only occurrence intervals. Moreover, time discretization rules are used in R-TDC to inspect a partial schedule in order to define a wait duration. The aim is to maximize the duration to speed up strategy search while limiting loss of possible strategies. Lastly, in order to improve completeness, we augment R-TDC waits with the possibility of instantaneous reactive executions \textbf{during strategy execution}. These are requests made to the waiting controller agent to immediately execute some controllable timepoint(s) when it observes an uncontrollable timepoint occur. Associated waits remain uninterruptible during strategy search however, resulting in only bounded and not exact scheduling times of controllable timepoints which are executed in such fashion, as shown in Figure~\ref{fig:dctypes}.

\begin{subdefinition}[R-TDC strategy structure]
A R-TDC strategy is a finite tree. This tree is comprised of a list of nodes ($N_1$,$N_2$, ..., $N_{q-1}$, $N_q$). Each $N_i$ is of the form:

$$N_i = (N_j, O_{ji}, E_i, \langle s_i, e_i, R_i \rangle) $$

\noindent where:

\begin{itemize}
    \item $N_j$ is the parent node of $N_i$ in the tree.
    \item Time $s_i$ is the start time of the wait in node $N_i$.
    \item Time $e_i$ is the end time of the wait in node $N_i$.
    \item $O_{ji}$ is the list of uncontrollable timepoints assumed to occur during the wait in node $N_j$. There exist as many $O_{ji}$ as the number of combinations of uncontrollable timepoints that may occur during the wait in node $N_j$. Therefore, in a R-TDC strategy, node $N_j$ will have exactly the same number of children nodes to account for all possible outcomes of uncontrollable timepoints.
    \item $R_i$ is a mapping which can associate to any uncontrollable timepoint that may occur in the wait $\langle s_i, e_i\rangle$ a set of controllable timepoints to reactively execute by the agent during strategy execution. The associated wait remains uninterrupted during strategy search even if some uncontrollable timepoints are assumed to occur in the wait.
    \item $E_i$ is a set of controllable timepoints to schedule at $s_i$.
\end{itemize}


\noindent Each path from the root node of a R-TDC strategy to any leaf node satisfies the following properties:

\begin{itemize}
    \item It covers the time horizon entirely (each new wait starts at the same time as the end of the previous wait).
    \item It represents a unique outcome of the occurrence possibilities of uncontrollable timepoints. Each uncontrollable timepoint is bounded in an occurrence interval $\langle s_i,e_i \rangle $. All possible outcomes are included in the strategy.
    \item It assigns to every controllable timepoint a given time of scheduling (or time interval for those reactively executed in response to uncontrollable timepoints).
    \item All constraints are satisfied given the scheduling time, scheduling time intervals and occurrence intervals of all timepoints.
\end{itemize}
\end{subdefinition}


\noindent We explain next how a R-TDC strategy is executed.

\vspace{-0.1cm}

\paragraph{R-TDC Strategy Execution.} A R-TDC strategy is executed in the following way by a controller agent. The agent starts at the root R-TDC node. For each current node $N_i = (N_j, O_{ji}, E_i, \langle s_i, e_i, R_i \rangle)$, it executes at $t=s_i$ the timepoints in $E_i$, and waits from time $s_i$ to $e_i$ with the reactive strategy $R_i$, \ie if $R_i$ stipulates it, the agent will immediately execute some controllable timepoints in response to some uncontrollable timepoints that may occur during the wait, as soon as they do. At the end of the wait, the agent deduces from the list of uncontrollable timepoints that occurred which child $N_i'$ of node $N_i$ it transitioned to. It moves to $N_i'$ and repeats the same process. Those guidelines are followed recursively until all constraints are satisfied.

We give a simple example of a R-TDC strategy for a DTNU $\gamma$ in Figure~\ref{fig:easy-example}. DTNU $\gamma'$ on the other hand is an example of a DTNU which is DC and TDC but not R-TDC. More precisely, it shows a clear limitation of R-TDC: when a controllable timepoint $a$ absolutely has to be scheduled a set time after an uncontrollable timepoint $u$ occurs: $a - u \in [x,x], x \in \mathbb{R}^+$. This is impossible in R-TDC as occurrence time of $u$ can only be bounded during strategy search and not exact, because any wait interval, however small, in which $u$ is assumed to occur is bounded.


\section{Tree Search Preliminaries}
\label{tree-search-prelim}
We introduce here the tree search algorithm. 
The root of the search tree built by the algorithm is a DTNU, and other tree nodes are either sub-DTNUs or logical nodes (\textit{OR, AND}) which respectively represent decisions that can be made and how uncontrollable timepoints can unfold. At a given DTNU tree node, decisions such as scheduling a controllable timepoint at current time or waiting for a period of time develop children DTNU nodes for which these decisions are propagated to constraints. In this tree, only one timepoint can be scheduled per branch, rather than a set of timepoints, simply for compatibility reasons with the heuristic function used for guidance. The R-TDC controllability of a \textit{leaf} DTNU node, \ie a sub-DTNU for which all controllable timepoints have been scheduled and uncontrollable timepoints are assumed to have occurred in specific intervals, indicates whether or not this sub-DTNU has been solved at the end of the scheduling process. We also refer to the R-TDC controllability of a DTNU node in the search tree as its \textit{truth attribute}. Lastly, the search logically combines R-TDC controllability of children nodes to determine R-TDC controllability for parent nodes.

\begin{figure}[tb]
\centering
\includegraphics[scale=0.62]{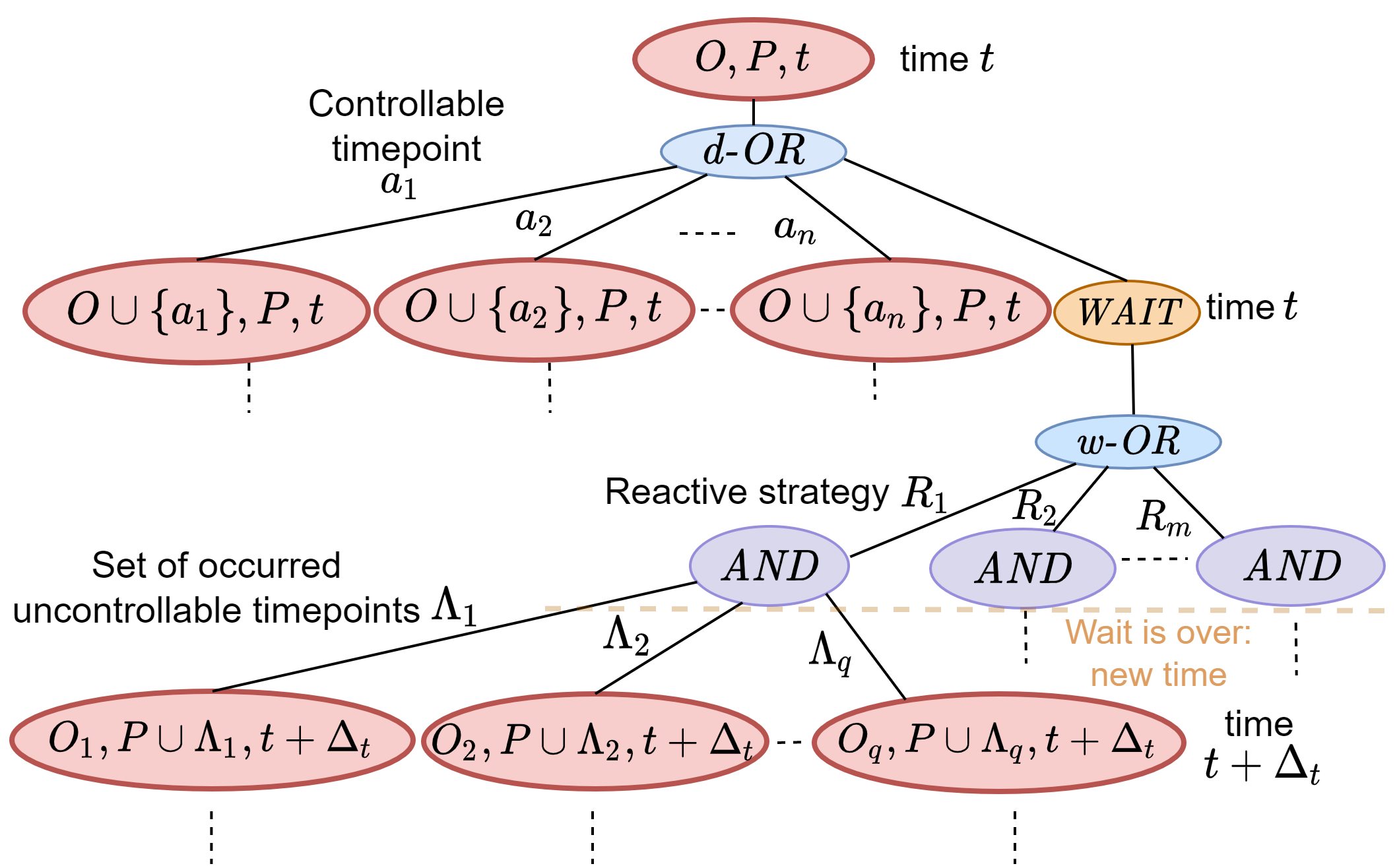}
\caption{\fontsize{10}{10} \normalfont \selectfont Structure of the search tree. 
Red nodes are DTNUs; $(O, P, t)$ is a DTNU for which $O$ is the set of controllable timepoints already scheduled, $P$ the set of uncontrollable timepoints that have occurred, and $t$ the time. Each branch $a_i$ refers to a controllable timepoint $a_i$, $R_i$ to a reactive strategy for the wait, and $\Lambda_i$ to a combination of uncontrollable timepoints.
}
\label{fig:ts-structure-fig}
\vspace{-0.5cm}
\end{figure}

Let $\Gamma = \{A,U,C,L\}$ be a DTNU.
The root node of the search tree is $\Gamma$. There are four different types of nodes in the tree and each node has a \textit{truth} attribute which is initialized to \textit{unknown} and can be set to either \true or \textit{false}. The different types of tree nodes are listed below and shown in Figure~\ref{fig:ts-structure-fig}.

\vspace{-0.15cm}

\paragraph{\textit{DTNU} nodes.} Any DTNU node other than the original problem $\Gamma$ corresponds to a sub-problem of $\Gamma$ at a given point in time $t$, for which some controllable timepoints may have already been scheduled in upper branches of the tree, some amount of time may have passed, and some uncontrollable timepoints are assumed to have occurred. A DTNU node is made of the same timepoints $A$ and $U$, constraints $C$ and contingency links $L$ as DTNU $\Gamma$. It also carries a schedule memory $S$ of what time controllable timepoints were scheduled during previous decisions in the tree, as well as the occurrence time intervals of uncontrollable timepoints assumed to have occurred. Lastly, the node also keeps track of the activation time intervals of \textit{activated} uncontrollable timepoints $B$ (uncontrollable timepoints that have been triggered by the scheduling of their associated controllable timepoint). The schedule memory $S$ is used to create an updated list of constraints $C'$ resulting from the propagation of the scheduling time or occurrence time interval of timepoints in constraints $C$. A non-terminal DTNU node, \ie a DTNU node for which all timepoints have not been scheduled, has exactly one child node: a \dor node.

\vspace{-0.15cm}
    
\paragraph{\textit{OR} nodes.} When a choice can be made at time $t$, this transition control is represented by an \textit{OR} node.
We distinguish two types of such nodes, \dor and \wor. For \dor nodes, the first type of choice is which controllable timepoint $a_i$ to schedule at current time. This leads to a DTNU node.
The other type of choice is to wait a period of time, which leads to a \wait node.
\wor nodes can be used to list reactive wait strategies, \ie to stipulate that some controllable timepoints will be set to be reactively executed to some uncontrollable timepoints in waits during strategy execution. 
The parent of a \wor node is therefore a \wait node and its children are \and nodes, described below.

\vspace{-0.15cm}

\paragraph{\wait nodes.} These nodes are used after a decision to wait a certain period of time $\Delta_t$. The parent of a \wait node is a \dor node. A \wait node has exactly one child: a \wor node, which has the purpose of exploring different reactive wait strategies. The uncertainty management related to uncontrollable timepoints is handled by \and nodes. 

\vspace{-0.15cm}
    
\paragraph{\and nodes.} Such nodes are used after a wait decision is taken and a reactive wait strategy is decided, represented respectively by a \wait and \wor node. Each child node of the \and node is a DTNU node at time $t + \Delta_t$, $t$ being the time before the wait and $\Delta_t$ the wait duration. Each child node represents an outcome of how uncontrollable timepoints may unfold and is built from the set of activated uncontrollable timepoints whose activation time interval overlaps the wait.  If there are $l$ activated uncontrollable timepoints, then there are at most $2^l$ \and node children, representing each element of the power set of activated uncontrollable timepoints.


Figure \ref{fig:ts-structure-fig} illustrates how a sub-problem of $\Gamma$, referred to as $DTNU_{O, P, t}$, is developed, where $O \subset A$ is the set of controllable timepoints that have already been scheduled, $P \subset U$ the set of uncontrollable timepoints which have occurred, and $t$ the time.
Moreover, two types of leaf nodes exist in the tree. The first type is a node $DTNU_{A, U, t}$ for which all controllable timepoints $a_i \in A$ have been scheduled and all uncontrollable timepoints  $u_i \in U$ have occurred. The second type is a node $DTNU_{A \setminus A', U, t}$ for which all uncontrollable timepoints $u_i \in U$ have occurred, but some controllable timepoints  $a_i \in A'$ have not been scheduled. The constraint satisfiability test of the former type of leaf node is straightforward: scheduling times and occurrence time intervals of all timepoints are propagated to constraints. For any timepoint whose occurrence time is only bounded in intervals and not exact, propagation is done in a way which assumes it could have occurred anywhere inside the interval to guarantee soundness. The leaf node's truth attribute is set to \true if all constraints are satisfied, \false otherwise. For the latter type, we propagate the occurrence time intervals of all uncontrollable timepoints as well as scheduling times of all scheduled controllable timepoints in the same way, and obtain an updated set of constraint $C'$. This leaf node, $DTNU_{A \setminus A', U, t}$, is therefore characterized as $\{A', \emptyset, C', \emptyset\}$ and is a DTN. We add the constraints $a'_i \geq t, \forall a'_i \in A'$ and use a mixed integer linear programming solver (\citet{cplex2009v12}) to solve the DTN. If a solution is found, the time values for each $a_i' \in A'$ are stored and the leaf node's truth value is set to \textit{true.} Otherwise, it is set to \textit{false.} After a truth value is assigned to the leaf node, the truth propagation function defined in \S~\ref{truthvalue} in the supplemental is called to logically infer truth value properties for parent nodes. A \true value reaching the root node of the tree means a R-TDC strategy has been found. The R-TDC strategy is a subtree of the search tree obtained by selecting recursively from the root, for each \dor and \wor nodes, the child with the \true attribute, and for each \and node, all children nodes (which are necessarily \textit{true}). A \false attribute reaching the root means there is no existing R-TDC strategy. As a result of the structure of the search tree which explores all possible outcomes of uncontrollable timepoints, and constraint propagation which enforces strict variable domain restrictions after an uncontrollable timepoint is bounded, the algorithm will always return sound strategies. Lastly, the search algorithm explores the tree in a depth-first manner.
We describe some simplifications made in the exploration in \S \ref{simplications} in the supplemental.

\section{Tree Search Characteristics}
\label{tree-search-chars}
We describe in this section how wait periods are calculated and how constraint propagation is performed. Moreover, we will designate as a \textit{conjunct} a constraint relationship of the form $v_i - v_j \in [x,y]$ or $v_i \in [x,y]$, where $v_i, v_j$ are timepoints and $x, y, \in \rm I\!R$. We refer to a constraint where several conjuncts are linked by $\lor$ operators as a \textit{disjunct}.

\subsection{Wait Action}
\label{wait}
When a wait decision of duration $\Delta_t$ is taken at time $t$ for a DTNU node, two categories of uncontrollable timepoints are considered to account for all transitional possibilities: \\

\vspace{-0.4cm}

\begin{itemize}
    \item $ Z = \{\zeta_1, \zeta_2, ..., \zeta_l\}$ is a set of timepoints that could either happen during the wait, or afterwards, \emph{i.e.} the end of the activation time interval for each $\zeta_i$ is greater than $t + \Delta_t$.
    \item $ H = \{\eta_1, \eta_2, ..., \eta_m\}$ is a set of timepoints that are certain to happen during the wait, \emph{i.e.} the end of the activation time interval for each $\eta_i$ is less than or equal to $t + \Delta_t$.
\end{itemize}{}


There are $q = 2^l$ number of different possible combinations (empty set included) $\Upsilon_1, \Upsilon_2, ..., \Upsilon_q$ of elements taken from $Z$. For each combination $\Upsilon_i$, the set $\Lambda_i = H \cup \Upsilon_i$ is created. The union $\bigcup\limits_{i=1}^{q} \Lambda_i$ refers to all possible combinations of uncontrollable timepoints which can occur by $t+ \Delta_t$. In Figure \ref{fig:ts-structure-fig}, for each \and node, the combination $\Lambda_i$ leads to a DTNU sub-problem $DTNU_{O_i, P \cup \Lambda_i, t + \Delta_t}$ for which the uncontrollable timepoints in $\Lambda_i$ are considered to have occurred between $t$ and $t + \Delta_t$ in the schedule memory $S$. In addition, any potential controllable timepoint $\phi$ planned to be instantly executed in a reactive wait strategy $R_i$ in response to an uncontrollable timepoint $u$ in $\Lambda_i$ will also be considered to have been scheduled between $t$ and $t + \Delta_t$ in $S$. The only exception is when checking constraint satisfiability for the conjunct $u - \phi \in [0,y]$ which required the reactive execution, for which we assume $\phi$ will be executed by the agent during strategy execution at the same time as $u$, thus the conjunct is considered satisfied.

\subsection{Wait Eligibility and Period}
\label{waitperiod}

The way wait durations are defined holds direct implications on the search space and the capability of the algorithm to find strategies. Longer waits make the search space smaller, but carry the risk of missing key moments where a decision is needed. On the other hand, smaller waits can make the search space too large to explore. We explain when the wait action is eligible, and how its duration is computed.

\vspace{-0.15cm}

\paragraph{Eligibility}
At least one of these two criteria has to be met for a \wait node to be added as child of a \dor node. \textbf{(1)} There is at least one activated uncontrollable timepoint for the parent DTNU node. \textbf{(2)} There is at least one conjunct of the form $v \in [x,y]$, where $v$ is a timepoint, in the constraints of the parent DTNU node. These criteria ensure that the search tree will not develop branches below \wait nodes when waiting is not relevant, \ie when a controllable timepoint necessarily needs to be scheduled. It also prevents the tree search from getting stuck in infinite \wait loop cycles.

\vspace{-0.15cm}

\paragraph{Wait Period} We define the wait duration $\Delta_t$ at a given \dor node by examining the updated constraint list $C'$ of the parent DTNU and the activation time intervals $B$ of its activated uncontrollable timepoints. Let $t$ be the current time for this DTNU node. Wait duration is defined by comparing $t$ to elements in $C'$ and $B$ to look for a minimum positive value defined by the next three rules. Each rule looks at the current partial schedule to identify \emph{'key milestones'} when actions should be taken, allowing to prefer longer waits when nothing is likely to happen before a long time, or shorter waits during critical moments. The purpose of the rules is to make it likely for there to be an existing R-TDC strategy when a DC one exists, while keeping the search space explored by the tree search algorithm as small as possible. \textbf{(1)} For each activated time interval $u \in [x,y] $ in $B$, we select $x - t$ or $y - t$, whichever is smaller and positive, and keep the smallest value $\delta_1$ found over all activated time intervals. This rule ensures the algorithm gets the opportunity to take a decision at the very beginning (or end) of a time frame in which an uncontrollable timepoint will occur. \textbf{(2)} For each conjunct $v \in [x,y] $ in $C'$, where $v$ is a timepoint, we select $x - t$ or $y - t$, whichever is smaller and positive, and keep the smallest value $\delta_2$ found over all conjuncts. This rule gives the algorithm the opportunity to act at the very beginning (or end) of a time frame in which it can satisfy a constraint requiring a timepoint to be in a specific interval. \textbf{(3)} We determine timepoints which need to be scheduled ahead of time by chaining constraints together. Intuitively, when a conjunct $v \in [x,y] $ is in $C'$, $v$ has to be scheduled when $t \in [x,y]$ to satisfy this conjunct. However, $v$ may be linked to other timepoints by constraints requiring them to happen before $v$. These timepoints may in turn be linked to yet other timepoints in the same way, and so on. Therefore, waiting until the time constraint window of $v$ may result in the algorithm actually over-waiting and being too late to tackle those constraint dependencies. The third rule consists in chaining backwards to identify potential timepoints starting this chain and potential time intervals in which they need to be scheduled. The following mechanism is used: for each conjunct $v \in [x,y] $ in $C'$ found in (2), we apply a recursive chain function to both $(v, x)$ and $(v, y)$. We detail how it is applied to $(v, x)$, the process being the same for $(v, y)$. Conjuncts of the form $v - v' \in [x', y'], x' \geq 0 $ in $C'$ are searched for. For each conjunct found, we add to a list two elements, $(v', x - x')$ and $(v', x - y')$. We select $x - x' - t$ or $x - y' - t$, whichever is smaller and positive, as potential minimum candidate. The chain function is called recursively on each element of the list. We keep the smallest candidate $\delta_3$. Figure \ref{dynamicwait} in the supplemental illustrates an application of this process. Finally, we set $\Delta_t = \min(\delta_1, \delta_2, \delta_3)$ as the wait duration. This duration is stored inside the \wait node.

\subsection{Reactive Executions during Waits}
\label{instant-scheduling}

Scheduling of a controllable timepoint may be necessary in some situations at the exact same time as when an uncontrollable timepoint occurs to satisfy a constraint. Therefore, different reactive wait strategies are considered and listed as children of a \wor node after a wait decision, before the start of the wait itself. If at any given DTNU node in the tree there is an activated uncontrollable timepoint $u$ with the potential to occur during the next wait and there is at least one unscheduled controllable timepoint $a$ such that a conjunct of the form $u - a\in [0,y], y \geq 0$ is present in the constraints, a reactive wait strategy is available that will set $a$ to be executed as soon as $u$ occurs during strategy execution. 

If there are $s$ controllable timepoints that may be set to be reactively executed, there are $2^s$ different reactive wait strategies $R_i$, each of which is embedded in an \and child of the \wor node. Let $\Phi = \{\phi_1, \phi_2, ..., \phi_s\} \subset A$ be the complete set of unscheduled controllable timepoints for which there are conjunct clauses $u - \phi_i \in [0,y]$. We denote as $R_1, R_2, ..., R_m$ all possible combinations of elements taken from $\Phi$, including the empty set. The child node $\textit{AND}_{R_i}$ of the \wor node resulting from the combination $R_i$ has a reactive wait strategy for which all controllable timepoints in $R_i$ will be immediately executed at the moment $u$ occurs during the wait, if it occurs.

\subsection{Constraint Propagation}
\label{tightbounds}

Decisions taken in the tree define when controllable timepoints are scheduled and also bear consequences on the occurrence time of uncontrollable timepoints. We explain here how these decisions are propagated into constraints, as well as the concept of `\textit{tight bound}'.

Let $C'$ be the list of updated constraints for a DTNU node $\boldsymbol{\psi}$ for which the parent node is $\boldsymbol{\omega}$. We distinguish two cases. Either $\boldsymbol{\omega}$ is a \dor node and $\boldsymbol{\psi}$ results from the scheduling of a controllable timepoint $a_i$, or $\boldsymbol{\omega}$ is an \and node and $\boldsymbol{\psi}$ results from a wait of $\Delta_t$ time units. In the first case, let $t$ be the scheduling time of $a_i$. The updated list $C'$ is built from the constraints of the parent DTNU of $\boldsymbol{\psi}$ in the tree. If a conjunct contains $a_i$ and is of the form $a_i \in [x,y]$, this conjunct is replaced with \textit{true} if $t \in [x,y]$, \textit{false} otherwise. If the conjunct is of the form $v_j - a_i \in [x,y]$, we replace the conjunct with $v_j \in [t + x, t + y]$. The other possibility is that $\boldsymbol{\psi}$ results from a wait of $\Delta_t$ time units at time $t$, with a reactive wait strategy $R$. In this case, the new time is $t + \Delta_t$ for $\boldsymbol{\psi}$. As a result of the wait, some uncontrollable timepoints $u_i \in \Lambda$ are assumed to have occurred, and some controllable timepoints $a_i \in A_R$ may be executed reactively during the wait. Let $v_i \in \Lambda \cup A_R$ be these timepoints occurring during the wait. The occurrence time of these timepoints is assumed to be in [$t$, $t+\Delta_t$]. For uncontrollable timepoints $u_i' \in \Lambda' \subset \Lambda$ for which the activation time ends at $t+\Delta_{t_i}'< t+\Delta_t$, and potential controllable timepoints $a_i'$ instantly reacting to these uncontrollable timepoints, the occurrence time is further reduced and considered to be in $[t, t+\Delta_{t_i}']$.

We define a concept of \textit{tight bound} to update constraints which restricts time intervals in order to account for all possible values $v_i$ can take between $t$ and $t+\Delta_t$. For all conjuncts $v_j - v_i \in [x,y]$, we replace the conjunct with $v_j \in [t + \Delta_t + x, t + y]$. Intuitively, this means that since $v_i$ can happen at the latest at $t + \Delta_t$, $v_j$ can not be allowed to happen before $t + \Delta_t + x$. Likewise, since $v_i$ can happen at the earliest at $t$,  $v_j$ can not be allowed to happen after $t + y$. Finally, if $t+ \Delta_t + x > t + y$, the conjunct is replaced with \false. Also, the process can be applied recursively in the event that $v_j$ is also a timepoint that occurred during the wait, in which case the conjunct would be replaced by \true or \textit{false}. In any case, any conjunct obtained of the form $a_j \in [x',y']$ is replaced with \false if $t + \Delta_t > y'$. Finally, if all conjuncts inside a disjunct are set to \false by this process, the constraint is violated and the DTNU is no longer satisfiable.
\section{Learning-based Heuristic}
\label{network}

We explain here how our learning model provides tree search guidance. Our MPNN architecture stems from \citet{gilmer2017neural}. It uses message passing rules enabling it to process graph-structured inputs. 
This architecture was originally designed for node classification in quantum chemistry and achieved state-of-the-art results on a molecular property prediction benchmark.
Here, we define a way of converting DTNUs into graph data. Then, we process the graph data with a fixed MPNN architecture and use the output to guide the tree search.

Let $\Gamma = \{A,U,C,L\}$ be a DTNU. We explain how we turn $\Gamma$ into a graph $\mathcal{G} = (\mathcal{K}, \mathcal{E})$. First, we convert all time values from absolute to relative by setting the current time for $\Gamma$ to $t=0$. We search all converted time intervals $[x_i,y_i]$ in $C$ and $L$ for the highest interval bound value $d_{max}$, \ie the farthest point in time. We normalize every time value in $C$ and $L$ by dividing them by $d_{max}$, yielding values between $0$ and $1$. Next, we convert each controllable timepoint $a \in A$ and uncontrollable timepoint $u \in U$ into graph nodes with corresponding \textit{controllable} or \textit{uncontrollable} node features. Time constraints in $C$ and contingency links in $L$ are expressed as edges between nodes with $10$ different edge distance classes ($0:[0,0.1)$, $1:[0.1,0.2)$, ...,  $9:[0.9,1]$). We also use additional edge features to account for edge types (constraint, disjunction, contingency link, direction sign for lower and upper bounds). Moreover, intermediary nodes are used with a distinct node feature in order to map possible disjunctions in constraints and contingency links. We 
add a \wait node with a distinct node feature which implicitly designates the act of waiting a period of time. Figure~\ref{fig-graph} in the supplemental shows an example of DTNU graph conversion.

The graph conversion of DTNU $\gamma$ contains three elements: the matrix of all node features $X_{\kappa}$, the adjacency matrix of the graph $X_{\epsilon}$ and the matrix of all edge features $X_{\rho}$. These features are processed by a fixed number of consecutive \textit{message passing} layers from \citet{gilmer2017neural} which make the MPNN. Each layer takes an input graph, consists of a phase during which messages are passed between nodes, and returns the same graph with new node features. Edge features remain the same. The overall process for a layer is as follows. For each node $\kappa_i$ in the input graph, a message passing phase creates new features for $\kappa_i$ from current features of neighboring nodes and edges. 
In detail, for each neighbor node $\kappa_j$, a small neural network 
(termed multi-layer perceptron, or MLP) takes as input the features of the edge connecting $\kappa_i$ and $\kappa_j$ and returns a matrix which is then multiplied by the features of $\kappa_j$ to obtain a feature vector. The sum of these vectors for the entire neighborhood defines the new features for $\kappa_i$. The output of the message passing layer consists of the graph updated with the new node features. 
In each message passing layer, the same MLP is used to process every node, so it can be applied to input graphs of any size, \ie the MPNN architecture can take as input DTNUs of any size. Moreover, each message passing layer uses a different MLP and can thus be trained to learn a different message passing scheme. Algorithm~\ref{messagePass} in the supplemental explains the workings of message passing.




Let $f$ be the function for our MPNN and $\theta$ its parameters. Function $f$ stacks $5$ message passing layers coupled with the $\mathrm{ReLU}(\cdot) = \max(0,\cdot)$ piece-wise activation function (\citet{glorot2011deep}) after each layer, except the last one. The first 4 layers have 32 abstract features per node, the last layer has 1 abstract feature per node. Each layer uses a trainable two-layer multi-layer perceptron (with 128 neurons in the hidden layer) for the message passing. Moreover, we add skip connections (\citet{he_2016}) to link each layer to the previous one. The $\texttt{sigmoid}$ function 
$\sigma(\cdot) = \frac{1}{1+\exp(-\cdot)}$ 
is used after the last layer to obtain a list of probabilities $\pi$ over all nodes in $\mathcal{G}$ : $f_{\theta}(X_{\kappa}, X_{\epsilon}, X_{\rho}) = \pi$. 
The probability of each node $\kappa$ in $\pi$ corresponds to the likelihood of transitioning into a R-TDC DTNU from the original DTNU $\Gamma$ by taking the action corresponding to $\kappa$. If $\kappa$ represents a controllable timepoint $a$ in $\Gamma$, its corresponding probability in $\pi$ is the likelihood of the sub-DTNU resulting from the scheduling of $a$ being R-TDC. If $\kappa$ represents a \wait decision, its probability refers to the likelihood of the \wait node having a \true attribute,
We call these two types of nodes \textit{active} nodes. Otherwise, if $\kappa$ is another type of node, its probability is not relevant to the problem and ignored. Our MPNN is trained on DTNUs generated and solved in \S~\ref{randomized-tree-search} in the supplemental only on active nodes by minimizing the binary cross-entropy loss:

\vspace{-0.3cm}

$$\frac{1}{m} \sum\limits_{i=1}^{m} \sum\limits_{j=1}^{q} - Y_{ij} \log (f_{\theta}(X_i)_j) - (1 - Y_{ij}) \log (1-f_{\theta}(X_i)_j)$$ 
Here $X_i = (X_{i_\kappa}, X_{i_\epsilon}, X_{i_\rho})$ is DTNU number $i$ among a training set of $m$ examples, $Y_{ij}$ is the R-TDC controllability (1 or 0) of active node number $j$ for DTNU number $i$. During training, we use batch normalization after each message passing layer. We add a dropout regularization layer with a \textit{keep rate} $0.9$ before the output layer to reduce overfitting. Training is done with the \textit{adagrad} optimizer from \citet{duchi2011adaptive} and an initial learning rate $10^{-4}$ on a dataset comprised of $30K$ instances generated as described in \S~\ref{randomized-tree-search} in the supplemental. We split the data into a training set comprised of $25K$ instances and a cross-validation set comprised of $5K$ instances on which we achieve $84\%$ accuracy. Lastly, the MPNN heuristic is used as follows in the tree search. Once a \dor node is reached, its parent DTNU node is converted into a graph and the MPNN is called upon the corresponding graph elements $X_{\kappa}, X_{\epsilon}, X_{\rho}$. Active nodes in output probabilities $\pi$ are then ordered by highest values first, and the search visits the corresponding children nodes in the suggested order, preferring children with higher likelihood of being R-TDC first.

\section{Experiments}

\begin{figure}[tb]
\centering
\includegraphics[scale=0.56]{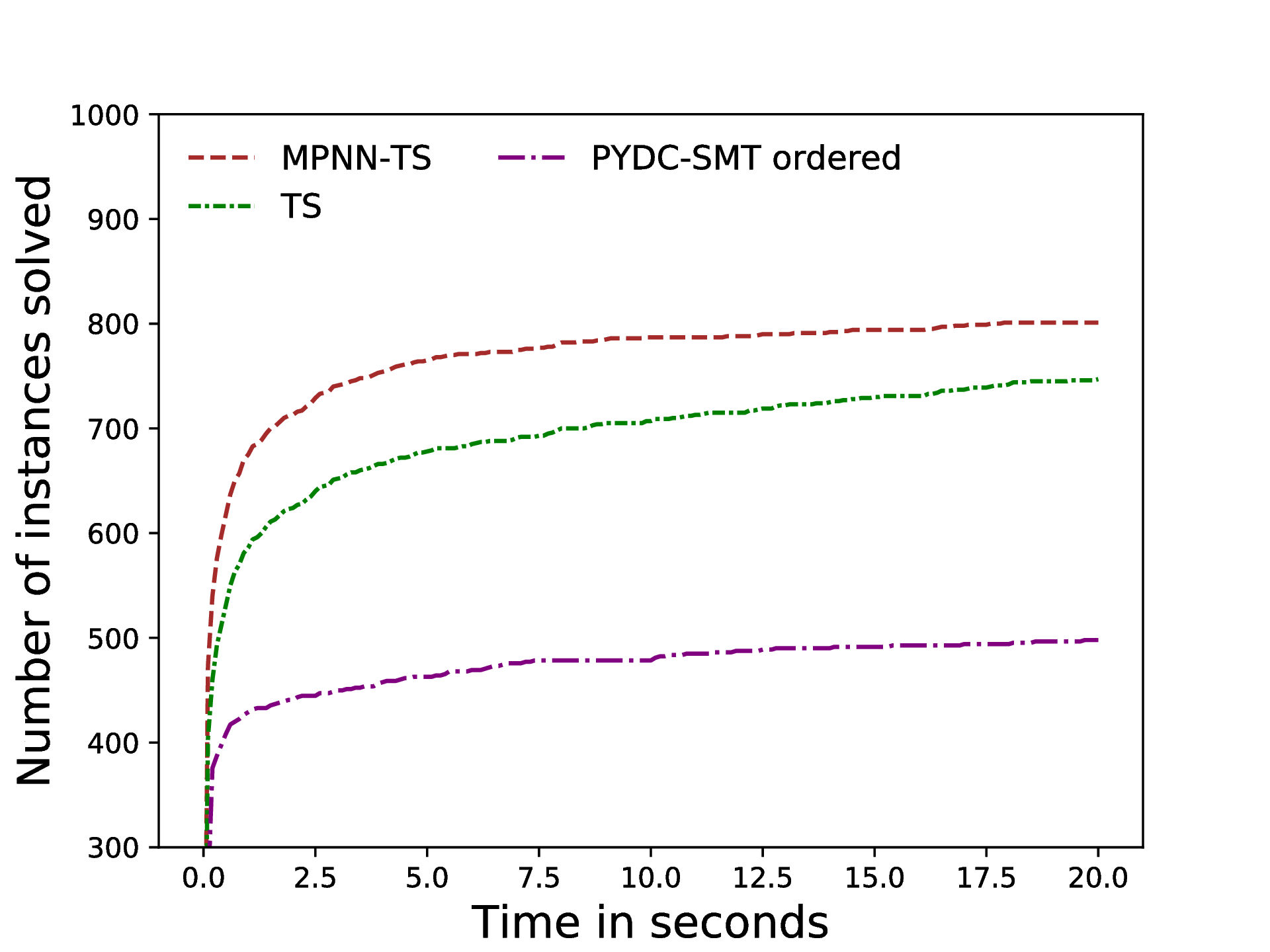}
\caption{\fontsize{10}{10} \normalfont \selectfont Experiments on \cite{cimatti2016dynamic}'s benchmark. The X-axis shows the allocated time (s) and the Y-axis the number of instances each solver can solve within the corresponding allocated time. Timeout is set to 20 seconds per instance.}
\label{aless-bench}
\vspace{-0.4cm}
\end{figure}


We evaluate the efficiency of the tree search and the effect of the MPNN's guidance. We also compare them to the state-of-the-art DC solver, PYDC-SMT-ordered, from \citet{cimatti2016dynamic} on a same computer. The tree search algorithm, trained MPNN and benchmarks are available \href{https://github.com/Liyue1d/DTNU-MPNN-TS}{here}. We use a laptop with the following specifications for experiments: $9^{th}$ gen. Intel Core i7, 16GB RAM and nvidia GTX 1660 Ti. R-TDC is a subset of DC and TDC: non-R-TDC controllability does not imply non-DC controllability. A R-TDC solver can thus be expected to offer better performance than a DC one while potentially being unable to find a strategy when a DC algorithm would. In this section, we refer to the tree search algorithm as TS and the tree search algorithm guided by the trained MPNN up to the $15^{th}$ (respectively $X^{th}$) \dor node depth-wise in the tree as MPNN-TS (respectively MPNN-TS-X).

First, we use the benchmark from \citet{cimatti2016dynamic} from which we remove DTNs and STNs. We compare TS, MPNN-TS and PYDC-SMT on the resulting benchmark which is comprised of $290$ DTNUs and $1042$ STNUs. Here, Limiting maximum depth use of the MPNN to $15$ offers a good trade off between guidance gain and cost of calling the MPNN. Results are given in Figure~\ref{aless-bench}. We observe TS solves roughly $50\%$ more problem instances than PYDC-SMT within the allocated time ($20$ seconds). In addition, TS solves $56\%$ of all instances while the remaining ones time out. Among solved instances, a strategy is found for $89\%$ and the remaining $11\%$ are proved non-R-TDC. On the other hand, PYDC-SMT solves $37\%$ of all instances. A strategy is found for $85\%$ of PYDC-SMT's solved instances, the remaining $15\%$ are proved non-DC. Finally, out of all instances PYDC-SMT solves, TS solves $97\%$ with the same conclusion, \ie R-TDC when DC and non-R-TDC when non-DC, highlighting the significant completeness retained by R-TDC. The use of the MPNN leads to an additional $+6\%$ problems solved. We argue this small increase is essentially due to the fact that most problems solved in the benchmark are small-sized problems with few timepoints which are solved quickly. Despite this fact, the MPNN still provides performance boost on a benchmark generated with another DTNU generator, suggesting the bias introduced by our DTNU generator remains limited and the MPNN is able to generalize to DTNUs created with a different approach.

\begin{figure}[t]
\centering
\includegraphics[scale=0.56]{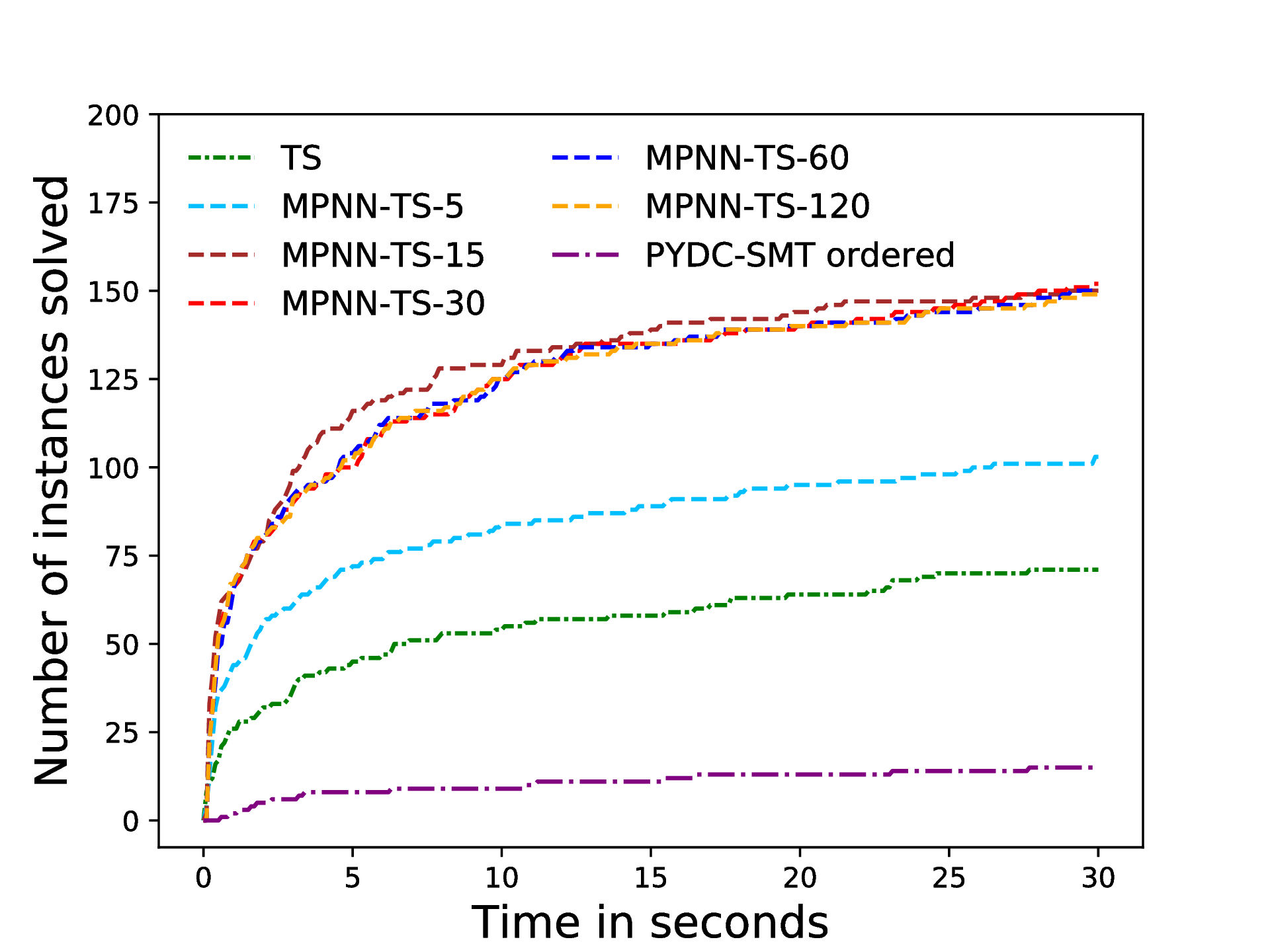}
\caption{\fontsize{10}{10} \normalfont \selectfont Experiments on benchmark $\boldsymbol{B_1}$. Axes are as in Figure \ref{aless-bench}. Timeout is set to 30 seconds per instance.}
\label{10-20}
\vspace{-0.5cm}
\end{figure}

For further evaluation of the MPNN, we create new benchmarks with the DTNU generator from \S~\ref{randomized-tree-search} (supplemental) with varying number of timepoints. These benchmarks have fewer 
quick to solve DTNUs and harder ones instead. Each benchmark contains 500 random DTNUs which have $1$ to $3$ uncontrollable timepoints. Moreover, each DTNU has 10 to 20 controllable timepoints in the $1^{st}$ benchmark $B_1$, 20 to 25 in the $2^{nd}$ benchmark $B_2$ and 25 to 30 in the last benchmark $B_3$. Each disjunct in the constraints of any DTNU contains up to 5 conjuncts. Experiments on $B_1$, $B_2$ and $B_3$ are respectively shown in Figure~\ref{10-20},~\ref{20-25-b} (in the supplemental) and \ref{25-30}. We note that for all three benchmarks no solver ever proves non-R-TDC or non-DC controllability before timing out due to the larger size of these problems.

PYDC-SMT performs poorly on $B_1$ and cannot solve any instance on $B_2$ and $B_3$. TS underperforms on $B_2$ and only solves 2 instances on $B_3$. However, we see a significantly higher gain from the use of the MPNN, varying with the maximum depth use. At best depth use, the gain is $+91\%$ instances solved for $B_1$, $+980\%$ for $B_2$ and $+1150\%$ for $B_3$. The more timepoints instances have, the more worthwhile MPNN guidance appears to be. Indeed, the optimal maximum depth use of the MPNN in the tree increases with the problem size: $15$ for $B_1$, $60$ for $B_2$ and $120$ for $B_3$. We argue this is due to the fact that more timepoints results in a wider search tree overall, including in deeper sections where MPNN use was not necessarily worth its cost for smaller problems. Furthermore, the MPNN is trained on randomly generated DTNUs which have 10 to 20 controllable timepoints. The promising gains shown by experiments on $B_2$ and $B_3$ suggest generalization of the MPNN to bigger problems than it is trained on.

The proposed tree search approach presents a good trade off between search completeness and effectiveness: almost all examples solved by PYDC-SMT in the benchmark of \citet{cimatti2016dynamic} are solved with the same conclusion, and many more which could not be solved are. Moreover, the R-TDC approach scales up better to problems with more timepoints, and the tree structure allows the use of learning-based heuristics. Although these heuristics are not key to solving problems of big scales, our experiments suggest they can still provide a high increase in efficiency.

\begin{figure}[t]
\centering
\includegraphics[scale=0.56]{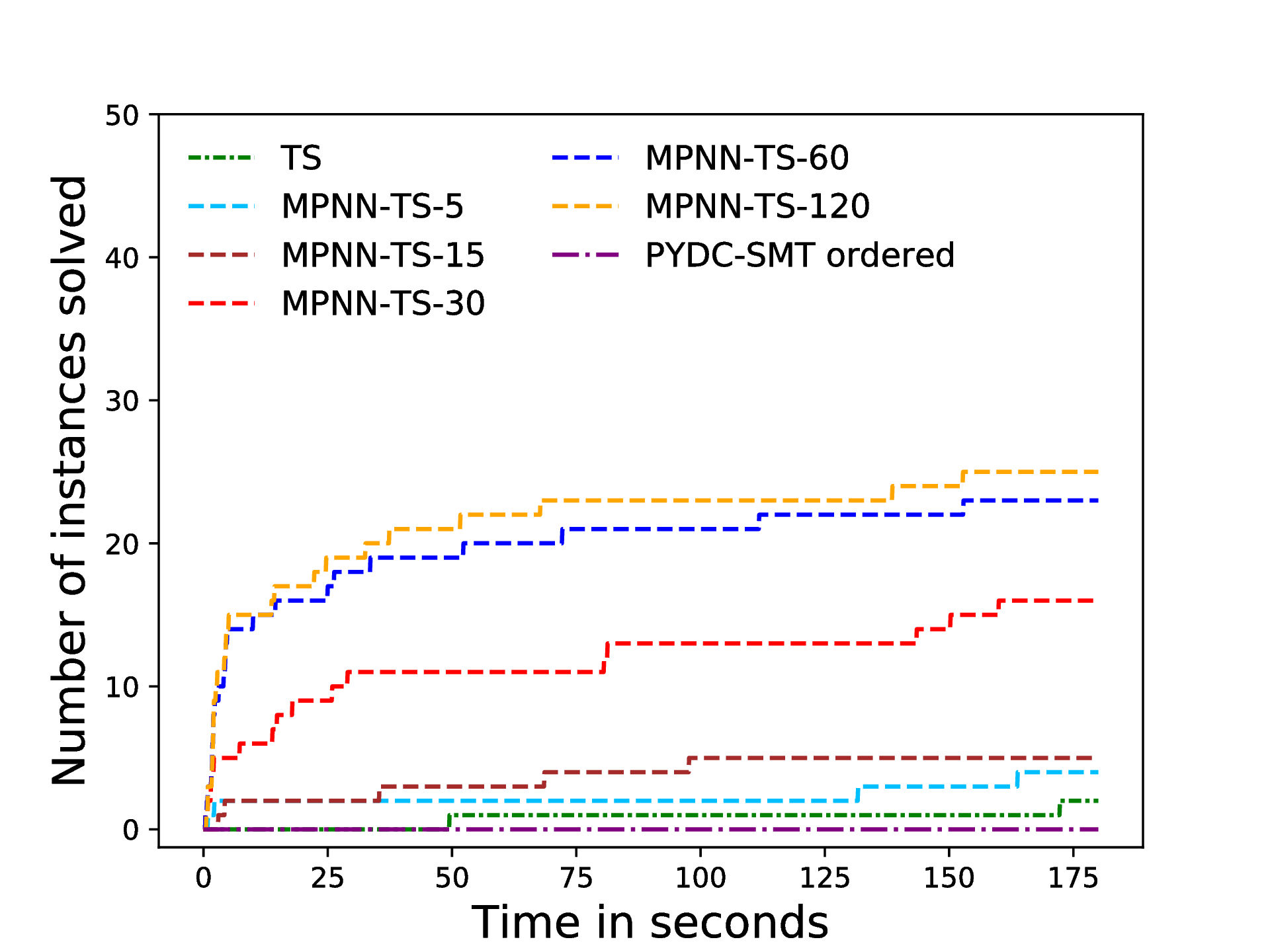}
  \caption{\fontsize{10}{10} \normalfont \selectfont Experiments on benchmark $\boldsymbol{B_3}$. Axes are as in Figure \ref{aless-bench}. Timeout is set to 180 seconds.}
\label{25-30}
\vspace{-0.5cm}
\end{figure}

\nocite{rusek2019unveiling} \nocite{chen2018deep} \nocite{xu2018experience} \nocite{Kool2018}
\nocite{wu2020comprehensive}
\nocite{ZhuwenLi2018}
\nocite{gasse2019exact}
\nocite{osanlou2019optimal}
\nocite{Ma2018}
\nocite{Katz2018}

\section{Conclusion}
We introduced new semantics for reactive scheduling: Time-based Dynamic Controllability (TDC) and a restricted subset of TDC, R-TDC. We presented a tree search approach for solving Disjunctive Temporal Networks with Uncertainty (DTNU) in R-TDC. Strategies are built by discretizing time and exploring different decisions which can be taken at different key points, as well as anticipating how uncontrollable timepoints can unfold.
We showed experimentally that R-TDC retains significant completeness, and enables the tree search approach to process DTNUs more efficiently than the state-of-the-art Dynamic Controllability (DC) solver does in DC. Lastly, we created MPNN-TS, a solver which combines the tree search with a heuristic function based on a graph neural network for guidance. The graph neural network enables steady improvements of the tree search on harder DTNU problems, notably on DTNUs of bigger size than those used for training the graph neural network.

\section*{Acknowledgements}

We would like to thank the reviewers whose feedback helped significantly improve the quality and clarity of this paper.

\nocite{osanlou2021constrained}
\nocite{li2021training}
\nocite{zhao2021distributed}
\nocite{park2021learning}
\nocite{zhang2020learning}
\nocite{wang2020learning}
\nocite{li2020graph}
\nocite{zhou2020variational}
\nocite{silver2020planning}
\nocite{hameed2020reinforcement}
\nocite{zhou2020reinforced}
\nocite{cappart2021combinatorial}
\nocite{wu2020comprehensive}
\nocite{sato2019approximation}
\nocite{prates2019learning}
\nocite{bhargava2019multiagent}
\nocite{zavatteri2019conditional}
\nocite{akmal2019quantifying}
\nocite{cimatti2018strong}
\nocite{cui2019dynamic}
\nocite{hu2020petri}
\nocite{li2021message}
\nocite{rusek2018message}
\nocite{hu2020collaborative}
\nocite{gasse2019exact}
\nocite{sievers2019deep}
\nocite{gama2021graph}
\nocite{nair2020solving}
\nocite{hu2020reinforcement}
\nocite{weng2020joint}
\nocite{garg2019size}
\nocite{sato2019approximation}
\nocite{lemos2019graph}
\nocite{drori2020learning}
\nocite{vesselinova2020learning}
\nocite{Guo2019SolvingCP}
\nocite{Wu2019ACS}
\nocite{Xu2019HowPA}
\nocite{Liu2020TowardsDG}
\nocite{Ying2019GNNExplainerGE}
\nocite{Carlo2019ConditionalST}
\nocite{Cui2015OptimisingBI}
\nocite{Santana2016PARISAP}
\nocite{Hunsberger2015EfficientEO}
\nocite{Nilsson2015EfficientPO}
\nocite{Cimatti2014UsingTG}
\nocite{Hunsberger2013MagicLI}
\nocite{Vidal2001DynamicSO}
\nocite{Lanz2015SimpleTN}
\nocite{Carlo2017AccessCT}
\nocite{Wang2019DeepGL}
\nocite{Rossi2020SIGNSI}
\nocite{Garg2020GeneralizationAR}
\nocite{Loukas2020WhatGN}
\nocite{osanlou2021learning}

\bibliography{prl}

\clearpage

\onecolumn

\section{Technical Appendix}

\subsection{Summary of Experiments on all Benchmarks}
\begin{figure*}[!htb]
    \renewcommand{\captionfont}{\small}
	\centering
	\begin{subfigure}[t]{0.47\textwidth}
		\centering
		\includegraphics[width=0.999\linewidth]{figures/aless300.jpg}
		\caption{Experiments on \cite{cimatti2016dynamic}'s benchmark. The X-axis represents the allocated time in seconds and the Y-axis the total number of instances that each solver can solve within the corresponding allocated time. Timeout is set to 20 seconds per instance.}\label{alless-b}		
	\end{subfigure}
	\quad
	\begin{subfigure}[t]{0.47\textwidth}
		\centering
		\includegraphics[width=0.999\linewidth]{figures/10-20.jpg}
		\caption{Experiments on benchmark $B_1$. Axes are the same as in figure \ref{alless-b}. Timeout is set to 30 seconds per instance.}\label{10-20-b}
	\end{subfigure}
\newline
	\centering
	\begin{subfigure}[t]{0.47\textwidth}
		\centering
		\includegraphics[width=0.999\linewidth]{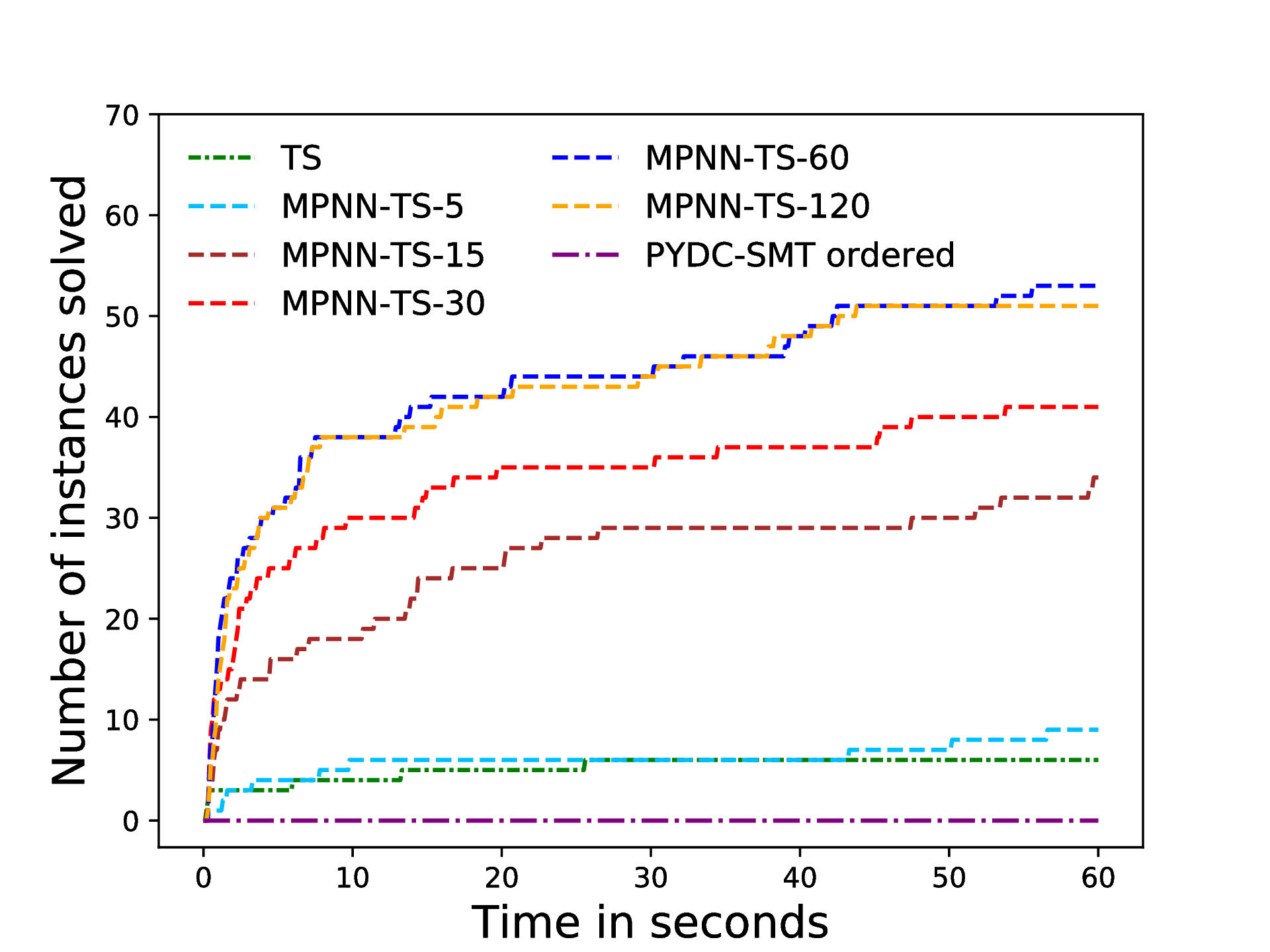}
		\caption{Experiments on benchmark $B_2$. Axes are the same as in figure \ref{alless-b}. Timeout is set to 60 seconds per instance.}
		\label{20-25-b}
	\end{subfigure}
	\quad
	\begin{subfigure}[t]{0.47\textwidth}
		\centering
		\includegraphics[width=0.999\linewidth]{figures/25-30.jpg}
		\caption{Experiments on benchmark $B_3$. Axes are the same as in figure \ref{alless-b}. Timeout is set to 180 seconds per instance.}
		\label{25-30-b}
	\end{subfigure}
	\caption{\textbf{Summary of experiments on benchmarks}}\label{fig-summary}
\end{figure*}

The plot for benchmark $B_2$, not provided in the paper, is given in Figure \ref{20-25-b}. 

\bigskip

\clearpage

\twocolumn

\bigskip




\subsection{Simplified Example}

\begin{figure}[H]
\renewcommand{\captionfont}{\small}
\centering
\includegraphics[scale=0.73]{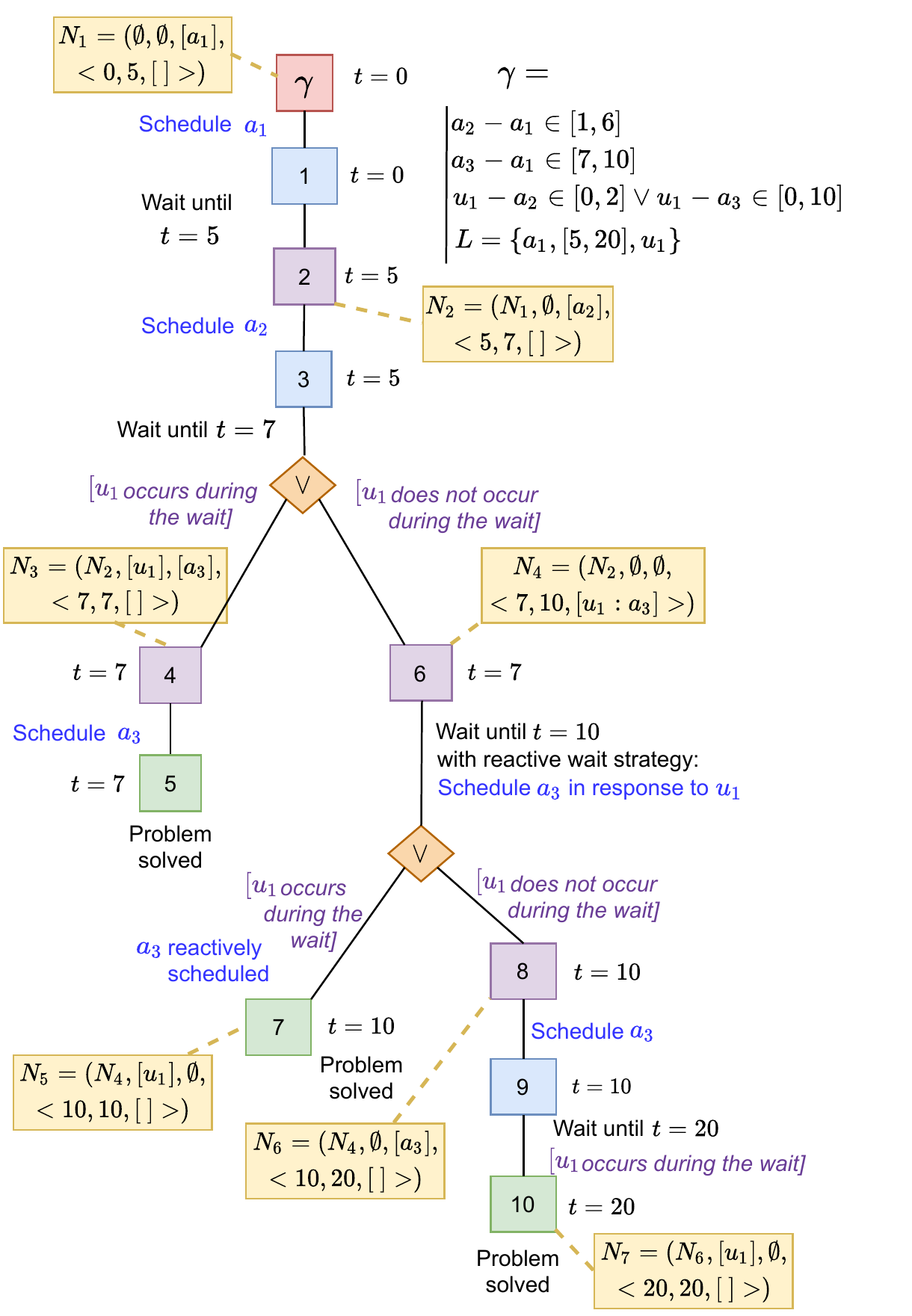}
\caption{\textbf{Detailed R-TDC strategy of a DTNU $\Gamma$.} The red square $\gamma$ is the original DTNU. Other squares are sub-DTNUs, except the $\lor$ signs which list transitional possibilities. Nodes $N_i$ represent the R-TDC strategy nodes as explained in the definition section.}
\label{alessandroexample}
\end{figure}

Figure \ref{alessandroexample} is a detailed R-TDC strategy of the example DTNU from \cite{cimatti2016dynamic}.

\subsection{Truth Value Propagation}
\label{truthvalue}
We describe how truth attributes of nodes are related to each other. The truth attribute of a tree node represents its R-TDC controllability, and the relationships shared between nodes make it possible to define sound strategies. When a leaf node is assigned a truth attribute $\beta$, the tree search is momentarily stopped and $\beta$ is propagated onto upper parent nodes. To this end, a  parent node $\boldsymbol{\omega}$ is selected recursively and we distinguish the following cases:

\begin{itemize}
    \item The parent $\boldsymbol{\omega}$ is a DTNU or \wait node: $\boldsymbol{\omega}$ is assigned $\beta$.
    \item The parent $\boldsymbol{\omega}$ is a \dor or \wor node: If $\beta = true$,  then $\boldsymbol{\omega}$ is assigned \true. If $\beta = false$ and all children nodes of $\boldsymbol{\omega}$ have \false attributes, $\boldsymbol{\omega}$ is assigned \false. Otherwise, the propagation stops.
    \item The parent $\boldsymbol{\omega}$ is an \and node: If $\beta = false$,  then $\boldsymbol{\omega}$ is assigned \false. If $\beta = true$ and all children nodes of $\boldsymbol{\omega}$ have \true attributes, $\boldsymbol{\omega}$ is assigned \true. Otherwise, the propagation stops.
\end{itemize}

\noindent After the propagation finishes, the tree search algorithm resumes where it was stopped. Algorithm \ref{truthvaluealgo} describes this process.

\begin{algorithm}[H]
\small
\caption{Truth Value Propagation}\label{truthvaluealgo}
\begin{algorithmic}[1]
\Function{propagateTruth(TreeNode $\psi$)}{}
    \State $\omega \gets $ parent($\psi$) \Comment{$^{1}*$}
    \If {$\omega = null$}
        \State \textbf{return}
    \EndIf
    \If {isDTNU($\omega$) or isWAIT($\omega)$}  \Comment{$^{2}*$}
	 	\State $\omega .truth \gets \psi .truth$
	 	\State propagateTruth($\omega$)
	\ElsIf {isOR($\omega$)}   \Comment{$^{3}*$}
	    \If {$\psi .truth = True$}
	        \State $\omega .truth \gets True$
	 	    \State propagateTruth($\omega$)
	 	\Else 
	 	    \If{$\forall \sigma_i, \sigma_i.truth = False$} \Comment{$^{4}*$}
	 	        \State $\omega .truth \gets False$
	 	        \State propagateTruth($\omega$)
	 	    \EndIf
	 	\EndIf
	
	\ElsIf {isAND($\omega$)} \Comment{$^{5}*$}
	    \If {$\psi .truth = False$}
	        \State $\omega .truth \gets False$
	 	    \State propagateTruth($\omega$)
	 	\Else 
	 	    \If{$\forall \sigma_i, \sigma_i.truth = True$} \Comment{$^{4}*$}
	 	        \State $\omega .truth \gets True$
	 	        \State propagateTruth($\omega$)
	 	    \EndIf
	 	\EndIf
	\EndIf

\EndFunction

\end{algorithmic}
\footnotemark{$*$ parent($x$): Returns the parent node of $x$, $null$ if none.\\}
\footnotemark{$*$ isDTNU($x$): Returns \True if $x$ is a DTNU node, \False otherwise; isWait($x$): Returns \True if $x$ is a \textit{WAIT} node, \False otherwise.\\}
\footnotemark{$*$ isOR($x$): Returns \True if $x$ is an \dor or \wor node, \False otherwise.\\}
\footnotemark{$*$ $\sigma_i$: Child number $i$ of $\omega$. For a \dor or \wor node, in the case where $\psi$ is \false but not all other children of $\omega$ are \false the propagation stops. Likewise, for an \and node and in the case where $\psi$ is \true but not all other children of $\omega$ are \true, the propagation stops.\\}
\footnotemark{$*$ isAnd($x$): Returns \True if $x$ is an \textit{AND} node, \False otherwise.\\}
\end{algorithm}


\vspace{5cm}

\subsection{Tree Search Algorithm}

We give the simplified pseudocode for the tree search in Algorithm \ref{tsalgo}.

\begin{algorithm}[H]
\small
\caption{Tree Search}\label{tsalgo}
\begin{algorithmic}[1]

\Function{explore(TreeNode $\psi$)}{}
    
    \If {$parent(\psi).truth \neq unknown$}
        \State \textbf{return}
    \EndIf

    \If {isDTNU($\psi$)}
    
        \State updateConstraints($\psi$) \Comment{$^{6}*$}
    
        \If{IsLeaf($\psi$)} \Comment{$^{7}*$}
            \State propagateTruth($\psi$)
            \State \textbf{return}
        \EndIf
        
        \State Create \dor child $\psi'$
        \State explore($\psi'$)

    \EndIf
    
    \If {isOR($\psi$)}
        \State Create list of all children $\Psi'$ \Comment{$^{8}*$}
        \For{$\psi'\in \Psi' $}
            \State explore($\psi'$)
        \EndFor
    \EndIf

    \If {isAND($\psi$)}
        \State Create list of all children $\Psi'$ \Comment{$^{9}*$}
        \For{$\psi'\in \Psi' $}
            \State explore($\psi'$)
        \EndFor
    \EndIf
    
    \If {isWAIT($\psi$)}
        \State create \wor child $\psi'$
        \State explore ($\psi'$)
    \EndIf
    
\EndFunction

\Function{main(DTNU $\gamma$)}{}
    \State explore($\gamma$)
    \If {$\gamma . truth = True$}
        \State \textbf{return} $True$
    \Else
        \State \textbf{return} $False$
    \EndIf
\EndFunction

\end{algorithmic}
\footnotemark{$*$  updateConstraints($x$): Updates the constraints of DTNU node $x$.}

\footnotemark{$*$ isLeaf($x$): Sets the truth value of $x$ to \true and returns \true if all constraints are satisfied. Sets the truth value to \false and returns \true if a constraint is violated. If no truth value can be inferred at this stage with the updated constraints, a second check is run to determine if all uncontrollable timepoints have occurred. If so, the corresponding DTN is solved, the truth value of $x$ is updated accordingly, and the function returns \true. Otherwise, no logical outcome can be inferred for the current state of the constraints because there remains at least one uncontrollable timepoint and this function returns \false.}

\footnotemark{$*$ If this is a \dor node, the list $\Psi'$ contains all the children DTNU nodes resulting from either the decision of scheduling a controllable timepoint, or the \wait node resulting from a wait if available. If this is a \wor node,  $\Psi'$ contains all \textit{AND}$_{R_j}$ nodes, each of which possess a reactive wait strategy $R_j$ }

\footnotemark{$*$ Here, the list $\Psi'$ contains all DTNUs resulting from all possible combinations $\Lambda_1, \Lambda_2, ..., \Lambda_q$ of uncontrollable timepoints which have the potential to occur during the current wait.}

\end{algorithm}


\subsection{Soundness of Strategies}
We remind the reader that if a \textit{true} attribute reaches the root node of the search tree of a DTNU, a R-TDC strategy has been found and search terminates. The R-TDC strategy is a subtree of the search tree obtained by selecting recursively from the root, for each \dor and \wor nodes, the child with the \true attribute, and for each \and node, all children nodes (which are necessarily \textit{true}). Hence, all nodes, including leaf nodes, of this sub-tree have a \textit{true} attribute. Rules of constraint propagation are as defined in \S~\ref{tightbounds}: when a controllable timepoint is scheduled at a \dor node, its exact execution time is incorporated to the partial schedule; when an uncontrollable timepoint is assumed to have occurred from $t$ to $t + \Delta_t$ during a wait, it is considered in the partial schedule that this timepoint has occurred at all possible times in the interval $[t, t + \Delta_t]$ through the concept of \textit{tight bound}. Furthermore, the structure of the sub-tree guarantees coverage of the entire time horizon and includes all possible outcomes of uncontrollable timepoints given wait durations.

\begin{lemma}
A R-TDC strategy found by the algorithm is sound and guarantees satisfiability of constraints.
\end{lemma}
\begin{proof}
  Let us assume there is a controllable timepoint $a$ in one of the leaf nodes with a schedule $t$ that causes unsatisfiability of constraints. This would cause the leaf node to have a \textit{false} attribute as a result of constraint propagation, which is contradictory. Let us now suppose there is an uncontrollable timepoint $u$ in one of the leaf nodes which is assumed to have occurred between $t$ and $t + \Delta_t$, and for which an occurrence time at $t' \in [t, t + \Delta_t]$ causes constraints of the leaf node to be violated. This also implies that propagating the interval $u \in [t, t + \Delta_t]$ in the constraints through the concept of tight bound results in the leaf node having a \textit{false} attribute since there is a time value $t' \in [t, t + \Delta_t]$ which violates constraints. This is contradictory as well.

\end{proof}

\subsection{Wait Period}

\begin{figure}[tb]
\renewcommand{\captionfont}{\small}
\centering
\includegraphics[scale=0.79]{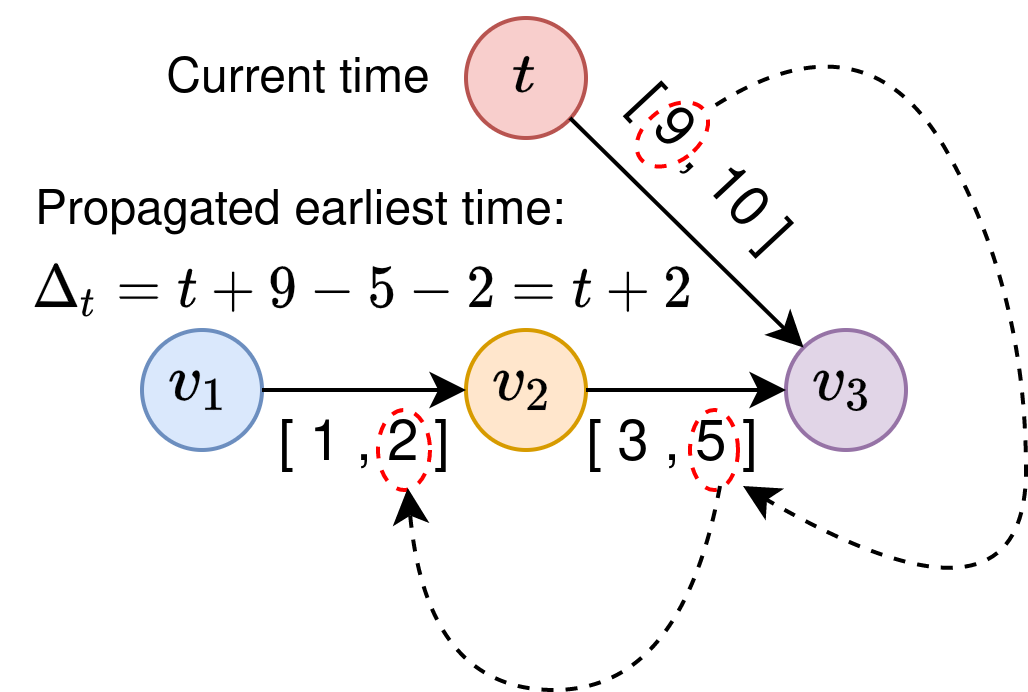}
\caption{\textbf{Application of the $3^{rd}$ rule to determine a wait duration.} Current time is $t$. Variables $v_1$, $v_2$ and $v_3$ are timepoints. Here, $v_2$ is constrained to be scheduled in the time interval $[1,2]$ after $v_1$, $v_3$ in $[3,5]$ after $v_2$ as well as in $[t+9, t+10]$. The rule suggests not to wait longer than $2$ units of time at $t$: scheduling $v_1$ at $t+2$, followed by scheduling $v_2$ at $t+4$ opens a window of opportunity for $v_3$ to be scheduled at $t+9$.}
\label{dynamicwait}
\vspace{-3mm}
\end{figure}

Figure \ref{dynamicwait} gives an example of the third rule used to compute a wait duration.

\subsection{Optimization Rules}
\label{searchoptimizations}
\label{simplications}
The following rules are added to make branch cuts.

\paragraph{\textit{Constraint Check.}} When a DTNU node is explored and the updated list of constraints $C'$ is built according to \S \ref{tightbounds}, if a disjunct is found to be \false, $C'$ will no longer be satisfiable. All the subtree which can be developed from the DTNU will only have leaf nodes for which this is the case as well. The search algorithm will not develop this subtree.

\vspace{-1mm}
    
\paragraph{\textit{Symmetrical subtrees.}} Some situations can lead to the development of the exact same subtrees. A trivial example, for a given DTNU node at a time $t$, is the order in which a given combination of controllable timepoints $a_1, a_2, ..., a_k$ is taken before taking a wait decision. Regardless of what order these timepoints are explored in the tree before moving to a \wait node, they will be considered scheduled at time $t$. When taking a wait decision, it is thus checked that all preceding controllable timepoints scheduled before the previous wait are a combination of timepoints that has not been tested yet.
    
\vspace{-1mm}    

\paragraph{\textit{Truth Checks.}}  Before exploring a new node for which the truth attribute is set to \textit{unknown}, the truth attribute of the parent node is also checked. The node is only developed if the parent node's truth attribute is set to \textit{unknown}. In this manner, when children of a tree node are being explored (depth-first) and the exploration of a child node leads to the assignment of a truth value to the tree node, the remaining unexplored children can be left unexplored.

\subsection{Self-Supervised Learning}
\label{randomized-tree-search}


We leverage a learning-based heuristic to guide the tree search. A key component in learning-based methods is the annotated training data. We generate such data in automatic manner by using a DTNU generator to create random DTNU problems and solving them with a modified version of the tree search. We store results and use them for training the MPNN. We detail here our data generation strategy.

We create DTNUs with a number of controllable timepoints ranging from 10 to 20 and uncontrollable timepoints ranging from 1 to 3. The generation process is the following. For interval bounds of constraint conjuncts or contingency links, we randomly generate real numbers within $[0,100]$. We restrict the number of conjuncts inside a disjunct to 5 at most. A random number $n_1 \in [10,20]$  of controllable timepoints and $n_2 \in [1,3]$ of uncontrollable timepoints are selected. Each uncontrollable timepoint is randomly linked to a different controllable timepoint with a contingency link. Next, we iterate over the list of timepoints, and for each timepoint $v_i$ not appearing in constraints or contingency links, we add in the constraints a disjunct for which at least one conjunct constrains $v_i$. The type of conjunct is selected randomly from either a \textit{distance} conjunct $v_i - v_j \in [x,y]$ or a \textit{bounded} conjunct $v_i\in [x,y]$. On the other hand, if $v_i$ was already present in the constraints or contingency links, we add a disjunct constraining $v_i$ with only a $20\%$ probability.

In order to solve these DTNUs, we modify the tree search as follows. 
For a DTNU $\Gamma$, the first \dor child node is developed as well as its children $\psi_1, \psi_2, ..., \psi_n \in \Psi$. The modified tree search explores each $\psi_i$ multiple times ($\nu$ times at most), each time with a timeout of $\tau$ seconds. 
We set $\nu = 25$ and $\tau = 3$. For each exploration of $\psi_i$, children nodes of any \dor node encountered in the corresponding subtree are explored randomly each time. If $\psi_i$ is proved to be either R-TDC or non-R-TDC during an exploration, the next explorations of the same child $\psi_i$ are called off and the truth attribute $\beta_i$ of $\psi_i$ is updated accordingly. The active node number $k$, corresponding to the decision leading to $\psi_i$ from DTNU $\Gamma$'s \dor node, is updated with the same value, \ie $Y_{k} = \beta_i$ ($1$ for \textit{true}, $0$ for \textit{false}). If every exploration times out, $\psi_i$ is assumed non-R-TDC and $Y_k$ is set to \textit{false}. Once each $\psi_i$ has been explored, the pair $\langle G(\Gamma), (Y_1, Y_2, ..., Y_n)\rangle$ is stored in the training set, where $G(\Gamma)$ is the graph conversion of $\Gamma$ 
described in \S\ref{network}. Data related to solved sub-DTNUs of $\Gamma$ are not stored in the training set as it was found to cause bias issues and overall decrease generalization in MPNN predictions.

The assumption of non-R-TDC controllability for children nodes for which all explorations time out is acceptable in the sense that the heuristic used is not admissible and does not need to be. The output of the MPNN is a probability for each child node of the \dor node, creating a preferential order of visit by highest probabilities first. Even in the event the suggested order first recommends visiting children nodes which will be found to be non-R-TDC, the algorithm will continue to explore the remaining children nodes until one is found to be R-TDC. Nevertheless, such a scenario rarely occurs in our experiments as the trained MPNN gives higher probabilities for children nodes for which explorations would tend to find a R-TDC strategy before timeout, and lower probabilities for ones where explorations would tend to result in a timeout.

\subsection{Message Passing Layer and DTNU to Graph Conversion}
\begin{figure}[tb]
\renewcommand{\captionfont}{\small}
\centering
\includegraphics[scale=0.65]{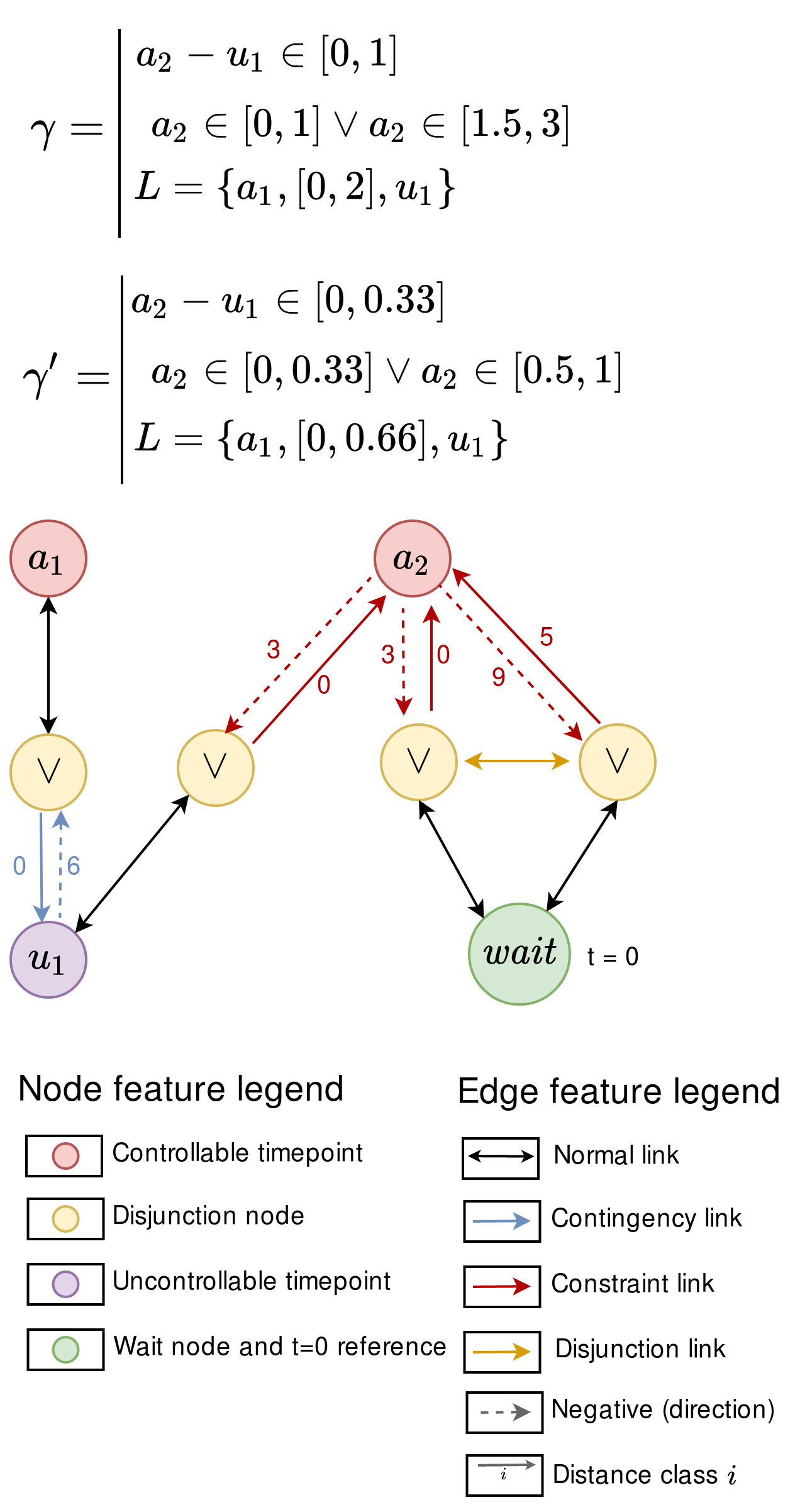}
\caption{\textbf{Conversion of a DTNU $\gamma$ into a graph.} $\gamma'$ is the normalized DTNU. Edge distances are expressed as distance classes. To distinguish between lower and upper bounds in intervals, we introduce an additional \textit{negative directional sign} feature.}
\label{fig-graph}
\vspace{-5mm}
\end{figure}

We use message passing layers that take as input a graph where nodes and edges possess features and return the graph with new node features. We detail pseudocode of a message passing layer applied to a graph $\mathcal{G=(K,E)}$ in Algorithm \ref{messagePass}. Additionally, we provide in Figure~\ref{fig-graph} an example of a DTNU converted into a graph.

\begin{algorithm}[thb]
\small
\caption{Message Passing Layer}\label{messagePass}
\begin{algorithmic}[1]
\Function{MsgPass(Graph $\langle \mathcal{(K,E)}, (H_\kappa, X_\epsilon, X_\rho)\rangle$)}{} \Comment{$^{10}*$}
    \State $H_\kappa'(\cdot,\cdot) \gets 0$ // \textit{Initialize new node features matrix}
    \ForAll{$\kappa_i \in \mathcal{K}$}
        \State $h_i' \gets 0$ // \textit{Initialize new features for $\kappa_i$}
        \ForAll{$\kappa_j \in \mathcal{K}$}
            \If{$X_{\epsilon}(\kappa_i,\kappa_j) = 1$}
                \State $\alpha \gets X_{\rho}(\kappa_i,\kappa_j)$
                \State $h \gets H_{\kappa}(\kappa_j,\cdot)$
                \State $h_i' \gets h_i' + MLP(\alpha)h$ \Comment{$^{11}*$}
            \EndIf
    
        \EndFor
        \State $H_\kappa'(\kappa_i, \cdot) \gets h_i'$ // \textit{Assign new features for $\kappa_i$}
    \EndFor
    \State \textbf{return} $\langle \mathcal{(K,E)}, (H_\kappa', X_\epsilon, X_\rho)\rangle$
\EndFunction
\end{algorithmic}
\footnotemark{$*$ $H_\kappa(\kappa_i, \cdot)$ returns a vector of current features for node $\kappa_i$; $X_\epsilon(\kappa_i, \kappa_j)$ returns $1$ if  $(\kappa_i, \kappa_j) \in \mathcal{E}$, $0$ otherwise; $X_\rho(\kappa_i, \kappa_j)$ returns a vector of current features for edge $(\kappa_i, \kappa_j)$.\\}
\footnotemark{$*$ MLP represents a multi-layer perceptron mapping input edge features to a matrix of dimension num-output-node-features x num-input-node-features. Moreover, $h$ is of dimension num-input-node-features x $1$. The matrix multiplication therefore results in a vector of size num-output-node-features. \\}

\end{algorithm}

\newpage

\subsection{Architecture Comparison}
\begin{figure}[h!]
\renewcommand{\captionfont}{\small}
\centering
\includegraphics[scale=0.45]{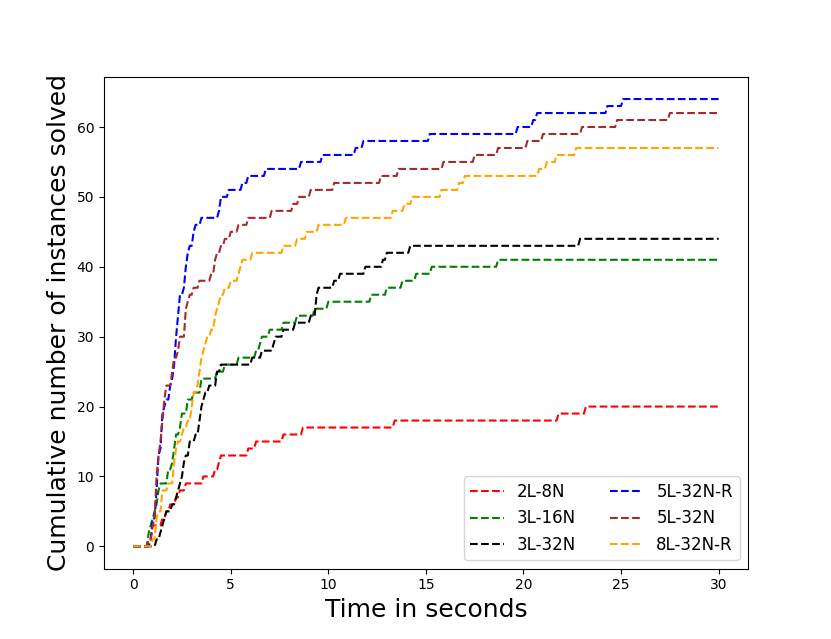}
\caption{\textbf{Comparison of different MPNN architectures.} 
Notation XL-YN refers to an MPNN with X layers and Y abstract node features per layer. The "-R" tag refers to the presence of residual layers.
Timeout is set to 30 seconds per DTNU instance.}
\vspace{50cm}
\label{fig-arch-comp}
\end{figure}

We study the impact of the design choices of the MPNN architecture on performance. To this end we compare different architectures of MPNN by varying depth and width (number of abstract node features per layer) and
train them on the training set created in \S \ref{randomized-tree-search}. 
We also 
assess the added value of residual skip connections to preceding layers. 
We create a benchmark of 400 DTNU instances, each of which has 20 to 25 controllable timepoints and up to 3 uncontrollable timepoints. We solve them using the tree search guided by each of these MPNN architectures. We limit the use of the MPNN architectures to a maximal depth of $50$ (\dor node-wise). Results are shown in Figure \ref{fig-arch-comp}. We note the smallest network is too small to learn efficiently and performs poorly. Three-layer networks perform better. 
Wider networks perform slightly better for the same depth, black network 32 vs. green network 16. 
Overall, medium-depth networks of 5 layers work best.
Residual connections lead to slight but steady gains.
Interestingly, deeper networks (8+ layers) display lower scores compared to more shallower variants (5 layers), suggesting depth performance saturation. The quantity of training data can however be a limiting factor: we assume the optimal architecture to be actually deeper.

\end{document}